\pdfoutput=1

\documentclass[11pt]{article}

\usepackage[final]{acl}

\usepackage{times}
\usepackage{latexsym}

\usepackage[T1]{fontenc}

\usepackage[utf8]{inputenc}

\usepackage{microtype}

\usepackage{inconsolata}

\usepackage{graphicx}

\usepackage{hyperref}
\usepackage{url}
\usepackage{enumitem}
\usepackage{tabularx}
\usepackage{amsfonts}
\usepackage{amsmath}
\usepackage{svg}
\usepackage{moresize}
\usepackage{caption}
\usepackage{subcaption}
\usepackage{enumitem}
\usepackage{xcolor}
\usepackage{multirow}
\usepackage{multicol}
\usepackage[para]{threeparttable}
\usepackage{booktabs}
\usepackage{array}
\usepackage{colortbl}   
\usepackage{xspace}
\usepackage[normalem]{ulem}
\useunder{\uline}{\ul}{}
\usepackage{pgffor}
\usepackage{longtable}
\usepackage{cleveref}
\usepackage{makecell}
\usepackage{csquotes}
\usepackage{mathtools}
\usepackage{kotex}

\usepackage{xparse}

\newcommand{\ea}{\ensuremath{e_1}}
\newcommand{\eb}{\ensuremath{e_2}}
\newcommand{\ec}{\ensuremath{e_3}}
\newcommand{\eea}{\ensuremath{E_1}}
\newcommand{\eeb}{\ensuremath{E_2}}
\newcommand{\eec}{\ensuremath{E_3}}

\newcommand{\raa}{\ensuremath{r_1}}
\newcommand{\rbb}{\ensuremath{r_2}}
\newcommand{\rcc}{\ensuremath{\rbb{\circ}\raa}}
\newcommand{\fa}{\ensuremath{\raa(\ea)}}
\newcommand{\fb}{\ensuremath{\rbb(\eb)}}
\newcommand{\fc}{\ensuremath{\rcc(\ea)}}
\newcommand{\qa}{\ensuremath{q(\fa)}}
\newcommand{\qb}{\ensuremath{q(\fb)}}
\newcommand{\qc}{\ensuremath{q(\fc)}}

\newcommand{\tc}{\ensuremath{(\qa, \qb, \qc, \eea, \eeb, \eec)}}

\newcommand{\wcolor}[1]{\textcolor{purple}{{#1}}}
\newcommand{\ccolor}[1]{\textcolor{teal}{{#1}}}

\newcommand{\unusable}{\textit{unusable}}
\newcommand{\guessable}{\textit{guessable}}
\newcommand{\nmodels}{41}
\newcommand{\enterc}{$\backslash$n}
\newcommand{\dataset}{\textsc{Socrates}}

\newcommand{\setify}[1]{$\{$#1$\}$}

\newcommand{\sourcee}{head}
\newcommand{\targete}{answer}

\newcommand{\textify}[1]{\textit{``#1''}}
\newcommand{\relationify}[1]{\texttt{#1}}

\makeatletter
\def\gcmidrule{\arrayrulecolor{gray!20}
    \noalign{\ifnum0=`}\fi
    \@ifnextchar[{\@gcmidrule}{\@gcmidrule[\cmidrulewidth]}}
\def\@gcmidrule[#1]{\@ifnextchar({\@@gcmidrule[#1]}{\@@gcmidrule[#1]()}}
\def\@@gcmidrule[#1](#2)#3{\@@@gcmidrule[#3]{#1}{#2}}
\def\@@@gcmidrule[#1-#2]#3#4{\global\@cmidla#1\relax
    \global\advance\@cmidla\m@ne
    \ifnum\@cmidla>0\global\let\@gtempa\@cmidrulea\else
    \global\let\@gtempa\@cmidruleb\fi
    \global\@cmidlb#2\relax
    \global\advance\@cmidlb-\@cmidla
    \global\@thisrulewidth=#3
    \@setrulekerning{#4}
    \ifnum\@lastruleclass=\z@\vskip \aboverulesep\fi
    \ifnum0=`{\fi}\@gtempa
    \noalign{\ifnum0=`}\fi\futurenonspacelet\@tempa\@xgcmidrule}
\def\@xgcmidrule{%
   \ifx\@tempa\gcmidrule
       \vskip-\@thisrulewidth
       \global\@lastruleclass=\@ne
   \else \ifx\@tempa\morecmidrules
       \vskip \cmidrulesep
       \global\@lastruleclass=\@ne\else
       \vskip \belowrulesep
       \global\@lastruleclass=\z@
   \fi\fi
   \ifnum0=`{\fi}
  \arrayrulecolor{black}}
\makeatother

\title{Do Large Language Models Perform Latent Multi-Hop Reasoning \\ \textit{without} Exploiting Shortcuts?}

\author{
\vspace{5px}
Sohee Yang\textsuperscript{1,2} \quad Nora Kassner\textsuperscript{1} \quad Elena Gribovskaya\textsuperscript{1} \quad Sebastian Riedel\textsuperscript{1,2$*$} \quad Mor Geva\textsuperscript{3,4$*$} \\ 
 \vspace{5px}
\textsuperscript{1}Google DeepMind \quad \textsuperscript{2}UCL \quad \textsuperscript{3}Google Research \quad \textsuperscript{4}Tel Aviv University \\  \vspace{2px}
\normalsize{\texttt{\{soheeyang,norakassner,egribovskaya,srriedel,pipek\}@google.com}}
}

\begin{document}
\maketitle
\def\thefootnote{*}\footnotetext{Corresponding authors.}\def\thefootnote{\arabic{footnote}}

\begin{abstract}
We evaluate how well Large Language Models (LLMs) latently recall and compose facts to answer multi-hop queries like \textify{In the year Scarlett Johansson was born, the Summer Olympics were hosted in the country of}.
One major challenge in such evaluation is that LLMs may have developed shortcuts by encountering the head entity \textify{Scarlett Johansson} and the answer entity \textify{United States} in the same training sequences or merely guess the answer based on frequency-based priors. To prevent shortcuts, we exclude test queries where the head and answer entities might have co-appeared during training. Through careful selection of relations and facts and systematic removal of cases where models might guess answers or exploit partial matches, we construct an evaluation dataset \dataset{} \textsc{(ShOrtCut-fRee lATent rEaSoning)}.
We observe that LLMs demonstrate promising latent multi-hop reasoning abilities without exploiting shortcuts, but only for certain types of queries. For queries requiring latent recall of countries as the intermediate answer, the best models achieve 80\% latent composability, but this drops to just 5\% for the recall of years. Comparisons with Chain-of-Thought highlight a significant gap between the ability of models to reason latently versus explicitly. Analysis reveals that latent representations of the intermediate answer are constructed more often in queries with higher latent composability, and shows the emergence of latent multi-hop reasoning during pretraining.\footnote{Our code and dataset are available at \url{https://github.com/google-deepmind/latent-multi-hop-reasoning} and \url{https://huggingface.co/datasets/soheeyang/SOCRATES}, respectively.}
\end{abstract}

\section{Introduction}

\begin{figure}[t!]
  \centering\includegraphics[width=0.5\textwidth]{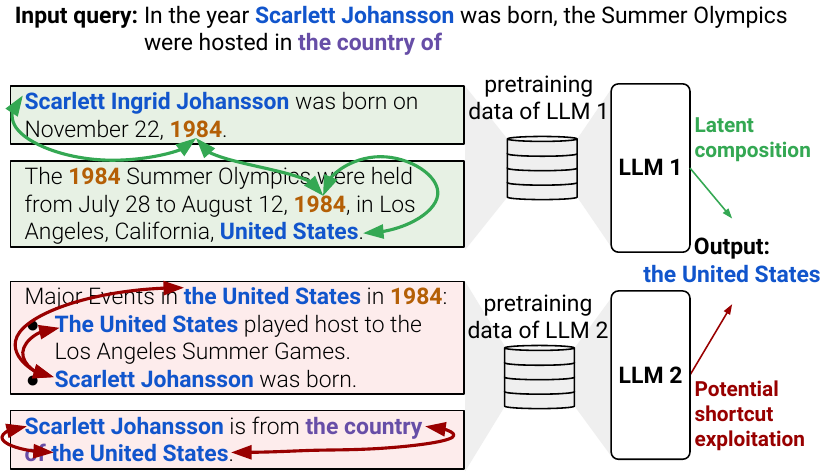}
\caption{Evaluation of latent multi-hop reasoning should exclude cases where LLMs can bypass the process of latently composing the single-hop facts by exploiting shortcuts. LLMs can develop shortcuts when they frequently encounter the head entity (\textify{Scarlett Johansson}) or the relation pattern in the query (\textify{the country of}) with the answer entity (\textify{United States}). We propose desiderata for shortcut-free evaluation of latent multi-hop reasoning ability.}
\label{fig:overview}
\end{figure}

Latent multi-hop reasoning in Large Language Models (LLMs), or latently recalling and composing single-hop facts to answer multi-hop queries like \textify{In the year Scarlett Johansson was born, the Summer Olympics were hosted in the country of}, has been of growing interest in recent years. First, this ability can be a measure towards better \textit{localization} and \textit{controllability} of factual knowledge in LLMs, as it can signal learning of a compressed representation of facts and their latent composition~\citep{Yang2024-ib}. This would provide more hope towards locate-then-edit or unlearn paradigm of LLMs~\citep{Meng2022-br, Hong2024-jn}. For instance, if complex facts are redundantly learned and recalled, edits with only single-hop facts would not propagate to the relevant multi-hop facts~\citep{Onoe2023-wm, Zhong2023-jl, Cohen2024-fm, Ju2024-do}. In addition, the ability to provide accurate answers without explicit Chain-of-Thought~(CoT) generation~\citep{Kojima2022-jt} could reduce inference costs. At the same time, whether LLMs can spontaneously develop latent reasoning abilities during pretraining is of interest from a safety perspective, as latent reasoning is less visible and hard to monitor given the opaque computations in LLMs~\citep{Berglund2023-tk, Treutlein2024-ob, Chan2024-vd}. Taken together, these incentives raise the question of \textit{How well do today's widely-used LLMs perform latent multi-hop reasoning over factual knowledge?}

If today's best models such as GPT-4o~\citep{openai2024gpt4o}, Claude 3.5 Sonnet~\citep{anthropic2024claude}, or Gemini 1.5 Pro~\citep{Gemini-Team2024-df} struggle to perform latent reasoning, then the current pretraining, instruction-tuning, and scaling paradigm of LLMs may be insufficient for robust development of latent multi-hop reasoning abilities. Thus, we would need to adopt changes in this paradigm or the architecture of models to enhance their ability to latently recall, compose, and manipulate parametric knowledge. However, if there are certain cases where today's LLMs show robust latent reasoning, we could further study these cases to find the underlying causes that make latent reasoning emerge during pretraining.

We evaluate latent multi-hop reasoning abilities by assessing models' performance in answering multi-hop queries. While prior works have suggested that pretrained LLMs develop this ability~\citep{Ofir-Press2023-dm,Yang2024-ib,Biran2024-gw,Li2024-oy}, they have not adequately addressed the possibility of models exploiting \textit{shortcuts}~\citep{Elazar2022-mi,Xu2022-po,Tang2023-so,Kang2023-hd,Ju2024-do}; our analysis shows that 85\% of the test queries from a previous dataset are prone to shortcuts.

Shortcuts from frequent co-occurrences of subject-object or relation-object in the training data can allow models to answer the multi-hop queries correctly without going through a true latent reasoning process. For instance, in a query about \textify{Scarlett Johansson} (i.e., the \emph{head entity}) where the answer entity is \textify{United States}, LLMs may simply learn a subject-object entity shortcut if these entities frequently co-occur in training~\citep{Elazar2022-mi, Kang2023-hd, Zhang2024-tt, Ju2024-do}. Similarly, LLMs can develop a relation-object shortcut from the frequency of \textify{United States} appearing as a country and guess the answer based on the frequency-based prior~\citep{Elazar2022-mi, Xu2022-po, Tang2023-so}.

To overcome this challenge, we outline desiderata for shortcut-free evaluation of latent multi-hop reasoning in LLMs, which we address through dataset construction and evaluation procedure. For the dataset, we only use test queries where the head and answer entities are estimated to never co-occur in pretraining sequences, thus minimizing subject-object shortcuts. We further minimize shortcuts by carefully selecting relation types and removing queries where the answers are easy to guess from a substring of the head entity.
For evaluation, we measure \emph{latent composability} as the rate of correct multi-hop answers when single-hop facts are known, without explicitly generating the intermediate answer (i.e., the \emph{bridge entity}). Additionally, we reduce relation-object shortcuts by excluding queries where the model may guess the answer without considering the head entity.

The main challenge in satisfying our desiderata is that most LLMs' pretraining data is inaccessible, making it impossible to directly check entity co-occurrences. To tackle this, we use a \emph{proxy corpus} of roughly 4.8B unique documents by utilizing six publicly available training corpora, selecting only test queries where the head and answer entities never co-occur. This approximation's effectiveness is validated by showing that strong latent composability for specific query types persists even when extending our entity co-occurrence check to the whole web via Google Search. Our resulting dataset, \dataset{} \textsc{(ShOrtCut-fRee lATent rEaSoning)}, consists of 7,232 pairs of single-hop and multi-hop queries of 17 types of relation compositions with 4 types of bridge entities.
Comparative experiments with a dataset constructed from the same data distribution but without careful fact selection, co-occurrence-based filtering, and rigorous evaluation show that latent composability can be overestimated without satisfying the desiderata.

Our results for \nmodels{} LLMs from 9 families reveal that there are successful cases of latent multi-hop reasoning, but the performance varies substantially according to the type of bridge entity that connects the facts. Notably, state-of-the-art models demonstrate strong latent composability of over 80\% when the bridge entity is a country. However, the number is only around 6\% for year-based queries, highlighting the importance of considering the distribution of relation composition types when evaluating LLMs' latent reasoning abilities. Models that know more single-hop facts tend to reason better latently, and the ability marginally improves with model scale. On the contrary, CoT composability effectively increases with the number of known facts and model size, with much higher and consistent performance across bridge entity types. Additional analysis shows that the latent representation of the bridge entity is clearly constructed more often for queries with higher latent composability, and reveals the emergence of latent multi-hop reasoning during pretraining.

In summary, our contributions are as follows:
\begin{itemize}[wide, labelindent=0pt, topsep=0.5pt, parsep=0.5pt, itemsep=0pt]
\item We present \dataset{} and evaluation procedure for latent multi-hop reasoning with minimal risk of shortcut exploitation, which is corroborated to be important through a comparative analysis.
\item We show that latent composability in LLMs significantly varies according to the bridge entity type.
\item We show that latent reasoning \textit{marginally} improves with the number of known single-hop facts and model scale and identify a significant gap between latent and CoT composability.
\item We present additional analysis results that help better understand LLMs' mechanisms for latent multi-hop reasoning.
\end{itemize}

\section{Related Work}

Studies have shown that LLMs' predictions often rely on shortcuts, shallow heuristics, and co-occurrence biases~\citep{Chen2019-kn,Jiang2019-ft,Geirhos2020-vw,Elazar2022-mi,Zhang2022-ot,Xu2022-po,Liu2023-tx,Tang2023-so,Kang2023-hd,Bachmann2024-jw,Ju2024-do}. For instance, \citet{Elazar2022-mi} have found that single-hop knowledge predictions can be influenced by subject-object co-occurrences or relation-object co-occurrences. Similarly, \citet{Kang2023-hd} and \citet{Zhang2024-tt} show that frequent co-occurrences can lead LLMs to favor high-frequency words over correct responses. Lastly, \citet{Ju2024-do} demonstrate that head-answer entity co-occurrence frequencies in multi-hop facts are correlated with factual shortcuts, which can cause failures in multi-hop knowledge editing.

Prior works on latent factual multi-hop reasoning have not fully addressed potential shortcuts~\citep{Ofir-Press2023-dm,Yang2024-ib,Biran2024-gw,Li2024-oy}. While \citet{Ofir-Press2023-dm} attempt to create multi-hop questions unlikely to appear in training, their approach relies on assumptions rather than co-occurrence statistics (leaving shortcuts from subject-object co-occurrences exploitable) and does not address shortcuts from relation-object co-occurrences. Indeed, we show that their dataset is prone to shortcuts, emphasizing the need for an explicit grounding to the co-occurrence statistics (see Section~\ref{sec:additional-analysis}). Similarly, \citet{Biran2024-gw} address relation-object shortcuts but overlook subject-object shortcuts.

Construction of a shortcut-free evaluation dataset of latent factual multi-hop reasoning ability of \emph{any pretrained LLM} presents a unique challenge, as the pretraining data of most of the widely used LLMs is not accessible, making it difficult to verify if certain single-hop facts or their composition appeared in the same training sequence. Our need to consider the knowledge LLMs learn during pretraining makes our work distinct from prior works that aim for shortcut-free evaluation on the tasks where the training dataset is fully accessible~\citep{Min2019-yi, Chen2019-kn, Ho2020-li, Trivedi2022-jp, Gregucci2024-wa}.

Other studies have attempted to circumvent these issues by fine-tuning LLMs on synthetic or counterfactual tasks to control for single-hop knowledge~\citep{Jiang2022-ku,Kassner2020-bw,Allen-Zhu2023-zc,Saparov2023-co,Hou2023-yl,Berglund2023-tk,Petty2024-wl,Treutlein2024-ob,Wang2024-tu}. However, these studies do not address our target question of how much latent multi-hop reasoning ability \emph{naturally} emerges in training. Moreover, finetuning may introduce side effects, such as hallucinations, reduced knowledge learning, and utilization efficiency~\citep{Yin2023-id, Kang2024-dk, Ghosal2024-fc, Gekhman2024-xc, Gottesman2024-fq}. Works on latent compositional reasoning with algorithmic or mathematical tasks~\citep{Dziri2023-vs,Chen2023-kp, Deng2024-gg} do not address our target question of latent multi-hop reasoning ability with factual knowledge, and may still suffer from data contamination that inflates performance~\citep{Zhang2024-ky}. \citet{Ko2024-bd} examine performance gaps between different numbers of reasoning hops but focus on more general reasoning capabilities rather than latent reasoning with factual knowledge.

\citet{Peng2024-jx} study the theoretical limitations of compositional abilities. Consistent with our findings, they prove that for high arity relations (like relations with year-type bridge entity in our work), multi-hop reasoning is more difficult, albeit for the specific case of Transformers~\citep{Vaswani2017-qc} with a single layer.

\section{Shortcut-Free Evaluation of Latent Multi-Hop Reasoning}\label{sec:desiderata}

\paragraph{Terms and Notations}
We represent single-hop facts as $\fa~=~\eb$ and $\fb~=~\ec$, and multi-hop facts as their composition $\fc = \ec$. In the aforementioned example, the entities \textify{Scarlett Johansson}, \textify{1984}, and \textify{United States} are respectively represented by \ea{}, \eb{}, \ec{}, and connected via relations \raa{} (\relationify{person-birthyear}), \rbb{} (\relationify{year-eventcountry}), and \rcc{} (\relationify{person-birthyear-eventcountry}). The set of aliases (names that the entity is also known as), of \ea{}, \eb{}, and \ec{} are represented as \eea{}, \eeb{}, and \eec{}, respectively. The answer set of the single-hop queries \qa{}, \qb{}, and multi-hop query \qc{} is \eeb{}, \eec{}, and \eec{}, respectively. For instance, the answer set corresponding to the query \textify{The year Scarlett Johansson was born in is} is $\{$\textify{1948}$\}$. We call each tuple $\tc{}$ a \textit{test case}, where \eb{} is a \textit{bridge entity} that connects the two facts, \ea{} is the \textit{\sourcee{} entity}, and \ec{} the \textit{answer} entity. Also, we call \textify{the year Scarlett Johansson was born} the \textit{descriptive mention} $\mu{}$ of the bridge entity.

\paragraph{Desideratum 1: Latent multi-hop reasoning}
We define the latent multi-hop reasoning ability of LLMs as the ability to latently recall and compose learned single-hop facts to answer multi-hop queries. We evaluate this ability by assessing a model's performance in answering multi-hop queries. For example, if a model learned the correct answer to \textit{``The year Scarlett Johansson was born in is''} and \textit{``In 1984, the Summer Olympics were hosted in the country of''}, we evaluate whether it recalls and composes these facts latently to correctly answer a multi-hop query like \textit{``In the year Scarlett Johansson was born, the Summer Olympics were hosted in the country of''}. Therefore, to exclusively evaluate \emph{latent} as opposed to \emph{explicit} reasoning, we require models to answer the query directly without generating intermediate results (e.g., \textit{``1984''}), e.g., without using Chain-of-Thought. Therefore, \textbf{the evaluation should exclude the cases where the model generates the intermediate answers before generating the final answer.}

\paragraph{Desideratum 2: Shortcut-free}
A model has a chance of exploiting shortcuts when it can correctly answer a multi-hop query by observing only part of the query (e.g., without \ea{} or \raa{} in the input). Shortcut exploitation is problematic for evaluating latent multi-hop reasoning abilities because it allows the model to bypass the need to latently recall and compose single-hop facts. Following \citet{Elazar2022-mi}, we consider two types of shortcuts: \textit{subject-object shortcuts}, where the model predicts objects that frequently co-occur with certain subjects or substrings of subjects, regardless of the relation semantics, and \textit{relation-object shortcuts}, where the model predicts objects that frequently appear with certain surface form text of a relation, regardless of the subject. Therefore, \textbf{the evaluation should exclude the queries prone to subject (or substring)-object shortcuts or relation (or paraphrase)-object shortcuts.}

\section{Evaluation Dataset}\label{sec:dataset}

In this section, we describe our dataset construction process that minimizes the chance of subject-object shortcut (satisfying Desideratum 2), resulting \dataset{} \textsc{(ShOrtCut-fRee lATent rEaSoning)}, a dataset for evaluation of shortcut-free evaluation of latent multi-hop reasoning.

\begin{table*}[ht!]
\centering
\resizebox{0.9\textwidth}{!}{\begin{tabular}{llrl}
\toprule
$\boldsymbol{\eb{}}$ \textbf{type} & \textbf{Relation Composition Type} & \textbf{Count} & \textbf{Example Multi-hop Query} \\
\midrule
city & person-birthcity-eventyear & 33 & $e_1$'s birth city hosted the Eurovision Song Contest in the year \\
\gcmidrule{1-4}
\multirow[c]{5}{*}{country} & university-locationcountry-anthem & 101 & The country where $e_1$ is in has the national anthem named \\
& university-locationcountry-isocode & 30 & The ISO 3166-1 numeric code of the country where $e_1$ is located is \\
& university-locationcountry-year & 7 & The founding year of the location country of $e_1$ is \\
& person-birthcountry-anthem & 22 & The name of the national anthem of $e_1$'s country of birth is \\
& person-birthcountry-isocode & 6 & The ISO 3166-1 numeric code used for the country where $e_1$ was born is \\
\gcmidrule{1-4}
\multirow[c]{2}{*}{university} & person-undergraduniversity-founder & 33 & The person who founded $e_1$'s undergrad university is named \\
& person-undergraduniversity-year & 25 & The year when the university where $e_1$ studied as an undergrad was founded is \\
\gcmidrule{1-4}
\multirow[c]{9}{*}{year} & person-birthyear-winner & 4,484 & The winner of the Nobel Prize in Literature in $e_1$'s birth year was \\
& city-eventyear-winner & 2 & In the year when the G7 Summit were hosted in $e_1$, the Nobel Prize in Chemistry was awarded to \\
& university-inceptionyear-winner & 632 & In the founding year of $e_1$, the Nobel Prize in Literature was awarded to \\
& university-inceptionyear-hostleader & 9 & The person who was the host leader of the G7 Summit in the founding year of $e_1$ is \\
& university-inceptionyear-eventcountry & 13 & In the year $e_1$ was founded, the host country of the G7 Summit was \\
& university-inceptionyear-eventcity & 62 & In the founding year of $e_1$, the host city of the G7 Summit was \\
& person-birthyear-eventcity & 1,389 & In the birth year of $e_1$, the Winter Olympics were hosted in the city of \\
& person-birthyear-hostleader & 260 & The leader of the host of the G7 Summit in $e_1$'s birth year is \\
& person-birthyear-eventcountry & 124 & The country where the Eurovision Song Contest took place in the birth year of $e_1$ is \\
\midrule
total & & 7,232 & \\
\bottomrule
\end{tabular}}
\caption{Dataset statistics and example queries of \dataset{}. The head entities are replaced with \ea{} to prevent potential data leakage. A more granular breakdown with the relation composition subtypes is in Appendix Table~\ref{tab:dataset_stats_subtype}.}
\label{tab:dataset_stats}
\end{table*}

\subsection{Dataset Construction}\label{sec:dataset_process}

We generate test cases from a knowledge graph $G$, where facts are represented as subject-relation-object triplets $\langle s, r, o \rangle$. Specifically, we collect pairs of facts $\fa = \eb$ and $\fb = \ec$ and their composition $\fc = \ec$ by considering pairs of triplets with a shared bridge entity, i.e., $\langle e_1, r_1, e_2 \rangle$, $\langle e_2, r_2, e_3 \rangle$. We choose Wikidata~\citep{Vrandecic2014-eu} as $G$.

\paragraph{Step 1: Selection of fact pairs}
We select single-hop facts that are likely well-known, but their composition is unlikely to naturally appear in general text corpora, to minimize the change of the model developing a shortcut between \ea{} and \ec{}. We observe that such cases typically occur when the set of possible options for \eb{} is large, there are numerous \ea{}'s that map to the same \eb{}, and the set of possible options for \ec{} is not too small (e.g., not \relationify{person-bloodtype}). This should lower the chance of the LLM getting the answer correct by mere guessing (Desideratum 2).

We exclude the following cases: (a) relation compositions where \ea{} and \ec{} are likely to be directly associated, e.g., \relationify{novel-maincharacter-creator}, (b) facts where the head and answer entities can be directly connected via popular single-hop relations other than the tested multi-hop relation, e.g., \relationify{person-birthyear-eventcountry( Scarlett Johansson) = United~States = person-birthcountry(Scarlett~Johansson)}, (c) queries with trivially inferrable bridge entities, e.g., \relationify{university-locationcountry (University~of~Washington)~=~United~States}, which could enable answer prediction via entity substring-based shortcuts, (d) relations where there are likely to be many entities with 1:n relation, such as \relationify{person-children}, and (e) single-hop facts with more than one answer (details in \S\ref{sec:impl12}).

\paragraph{Step 2: Test case construction}

We convert the selected fact pairs into a set of test cases $\{\tc\}$. To create the two single-hop queries \qa{}, \qb{} and their corresponding multi-hop query \qc{}, we follow the common practice of using diverse handcrafted natural language templates~\citep{Petroni2019-lb,Yang2024-ib}. For each relation, we use 16 templates (4 for each of the two single-hop queries) and randomly sample one template for each query, resulting in approximately 100K test cases. We construct the queries as incomplete sentences, instead of questions, so that the test query can be naturally completed by any pretrained model to derive the answer without finetuning.

\paragraph{Step 3: Test case filtering using training co-occurrence statistics}

We filter out cases where any aliases of the \sourcee{} entity \ea{} and \targete{} entity \ec{} co-occur in the same sequence that the evaluated LLM has seen during pretraining, preventing subject-object shortcuts. However, since pretraining sequences and/or corpora of most LLMs are often inaccessible, we approximate co-occurrences by checking document-level co-occurrences across a \textit{proxy corpus} of 4.8B unique documents (details in \S\ref{sec:impl3}). While this approximation cannot guarantee complete exclusion of co-occurring entities without access to exact pretraining corpora, we validate our approach using Google Search for web-wide co-occurrence verification (\S\ref{sec:search}).

\paragraph{The \dataset{} Dataset}
\dataset{} contains 7,232 test cases of 17 types of relation compositions connected by 4 types of bridge entities, as shown in Table~\ref{tab:dataset_stats}. Note that the distribution of relation compositions is imbalanced as \ea{} and \ec{} of some relation compositions frequently appear together in the same document and most of test cases are removed by the co-occurrence-based filtering.

\begin{figure*}[t!]
  \centering
  \begin{subfigure}[b]{.99\textwidth}
  \includegraphics[width=\textwidth]{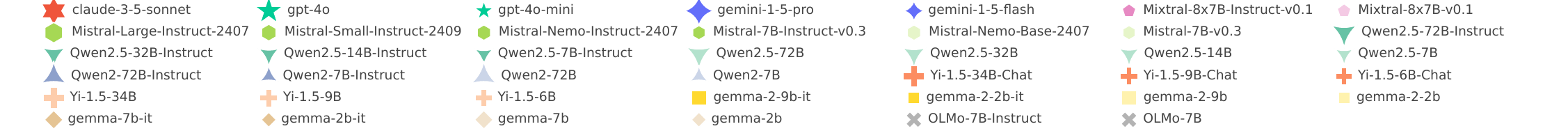}
  \label{fig:legend_models}
  \end{subfigure}
  \begin{subfigure}[b]{.49\textwidth}
  \includegraphics[width=\textwidth]{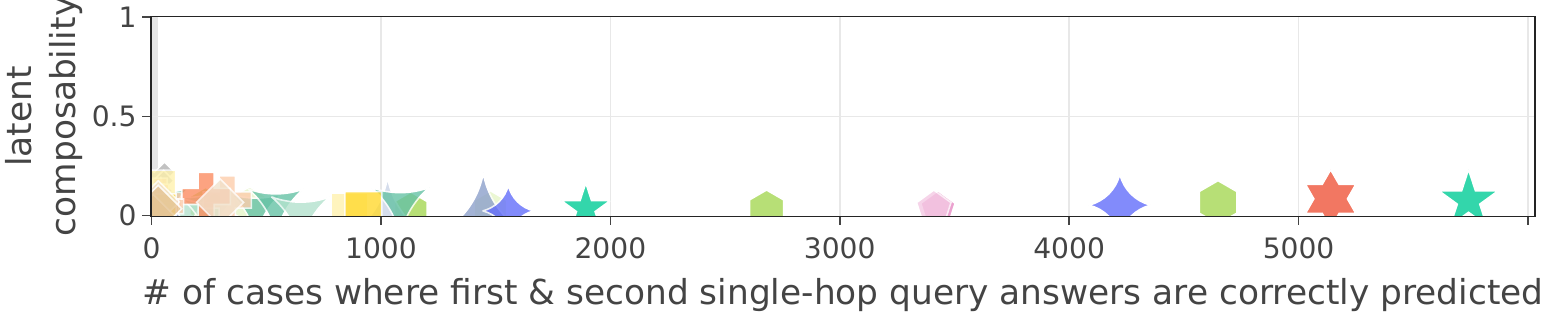}
  \caption{Latent composability. There are successful cases of latent multi-hop reasoning, although the overall percentage is low.}
  \label{fig:latent_composability}
  \end{subfigure}\hfill
  \begin{subfigure}[b]{.49\textwidth}
  \includegraphics[width=\textwidth]{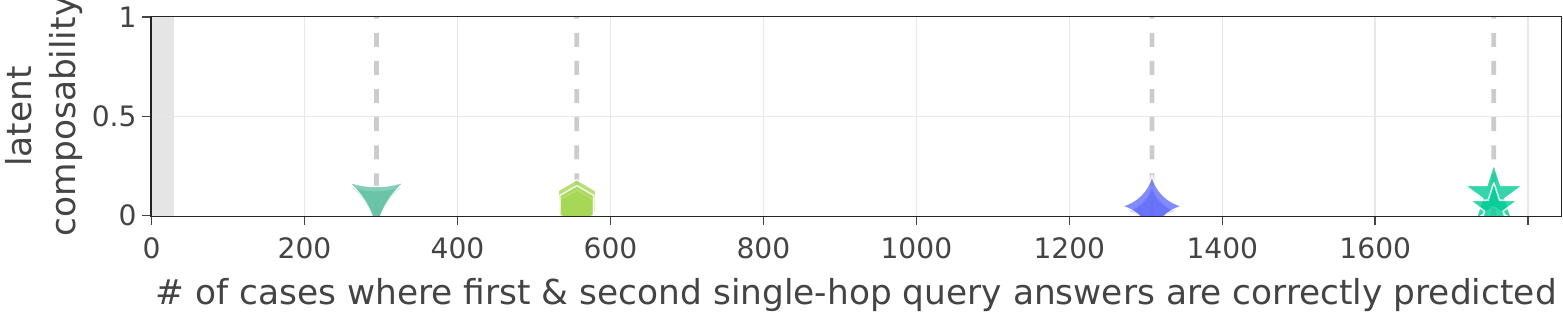}
  \caption{Effect of model scale on latent composability. Comparative latent composability slightly improves with model scale.}
  \label{fig:comparative_latent_composability}
  \end{subfigure}
  \begin{subfigure}[b]{.49\textwidth}
  \includegraphics[width=\textwidth]{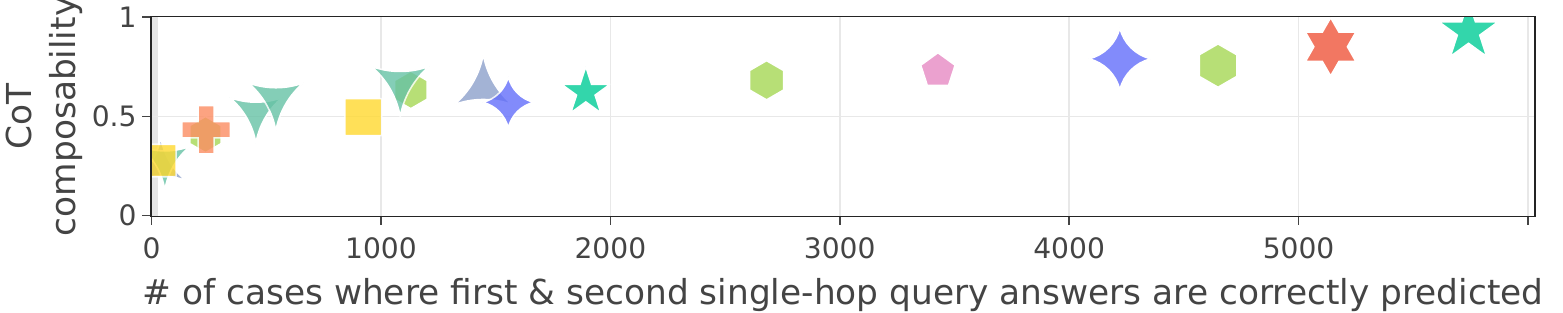}
  \caption{CoT composability. CoT composability is significantly higher than latent composability.\\}
  \label{fig:cot_composability}
  \end{subfigure}\hfill
  \begin{subfigure}[b]{.49\textwidth}
  \includegraphics[width=\textwidth]{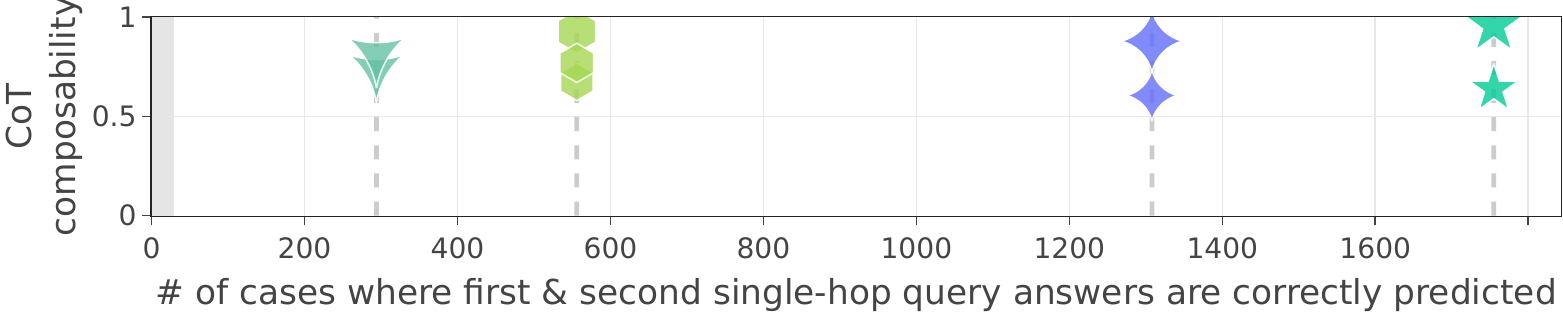}
  \caption{Effect of model scale on CoT composability. Comparative CoT composability increases much more effectively with model scale compared to latent composability.}
  \label{fig:comparative_cot_composability}
  \end{subfigure}
\caption{Latent \textbf{(upper row)} and CoT \textbf{(lower row)} composability on \dataset{}.}
\label{fig:composability_main}
\end{figure*}

\section{Evaluation Procedure}\label{sec:metric}

We introduce an evaluation procedure that satisfies part of Desideratum 2 by minimizing the chance of the model exploiting the relation-object shortcut and Desideratum 1 that the evaluation should exclude cases where the model performs explicit reasoning (\S\ref{subsec:per_query_prediction}). Then, we define our evaluation metric, latent composability (\S\ref{subsec:latent_composability}).

\subsection{Filtering Guessable and Unusable Queries}
\label{subsec:per_query_prediction}

\paragraph{Excluding \guessable{} cases}
Even when the prediction of the model for a multi-hop query is correct, there is still a chance that the LLM might have guessed the answer correctly by chance, meaning that Desideratum 2 is not satisfied. For instance, when the answer is a popular entity among the potential answer set (e.g., \textify{United States} as a country), the model might exploit a relation-object shortcut where the model makes the prediction based on the prior from the textual pattern of the relation (e.g., \textify{$\ldots$ is from the country of}).

To filter out the cases where it is indistinguishable whether models are guessing the answer, we adopt the method of \citet{Biran2024-gw} which checks the model's prediction for a set of \emph{ablated prompts} $Q_\emptyset = \{ q(\rcc(\emptyset)), q(\rbb(\emptyset)) \}$, where the specific information of \ea{} and \fa{} is ablated from the multi-hop query \qc{}, respectively (e.g., $\{$\textify{In the year the person was born, the Summer Olympics were hosted in the country of}, \textify{In the year, the Summer Olympics were hosted in the country of}$\}$). When the model answers the multi-hop query correctly, but also answers any of $q_\emptyset \in Q_\emptyset$ correctly, we exclude the test case from the evaluation. Namely, we detect and remove the cases where models may be exploiting a relation-object shortcut between \rcc{}/\rbb{} and \ec{}.

\paragraph{Excluding \unusable{} cases}
When the model correctly predicts the answer for a test multi-hop query but the LLM has just enumerated multiple potential answer candidates\footnote{\textify{A. United States B. Canada C. United Kingdom}} or the LLM has performed an explicit reasoning (e.g., \textify{1984, United States}), we view these test cases as \textit{\unusable{}} for the evaluation of latent multi-hop reasoning ability and exclude the test case from the evaluation, following Desideratum 1 (details in \S\ref{sec:cheat}).

\paragraph{Suppressing CoT for instruction-tuned LLMs}\label{sec:suppress_cot}
Since instruction-tuned LLMs tend to perform CoT-style reasoning by default\footnote{\textify{Scarlett Johansson was born in 1984, and the Summer Olympics that year were hosted by the United States.}}, we formulate the task as a fill-in-the-blank task using a CoT-suppressing instruction as described in \S\ref{sec:instruction}.\footnote{While it is possible to use few-shot learning to restrict the format of the answer \cite{Ofir-Press2023-dm}, we use instructions to avoid potential biases in the selection of the few-shot demonstrations~\citep{Zhang2022-nk}.}

\begin{figure*}[t!]
  \centering
  \begin{subfigure}[b]{\textwidth}
    \begin{minipage}[b]{.49\textwidth}
      \includegraphics[width=\textwidth]{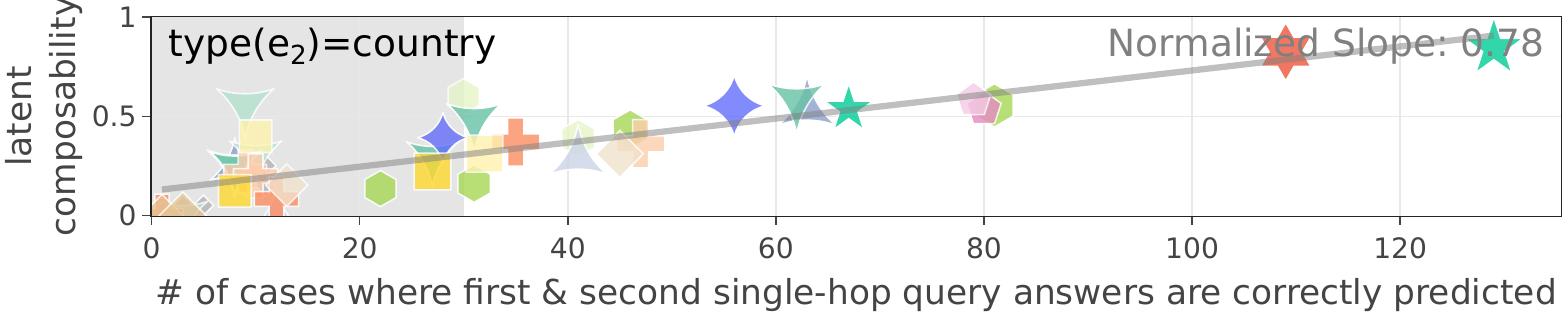}
    \end{minipage}\hfill
    \begin{minipage}[b]{.49\textwidth}
      \includegraphics[width=\textwidth]{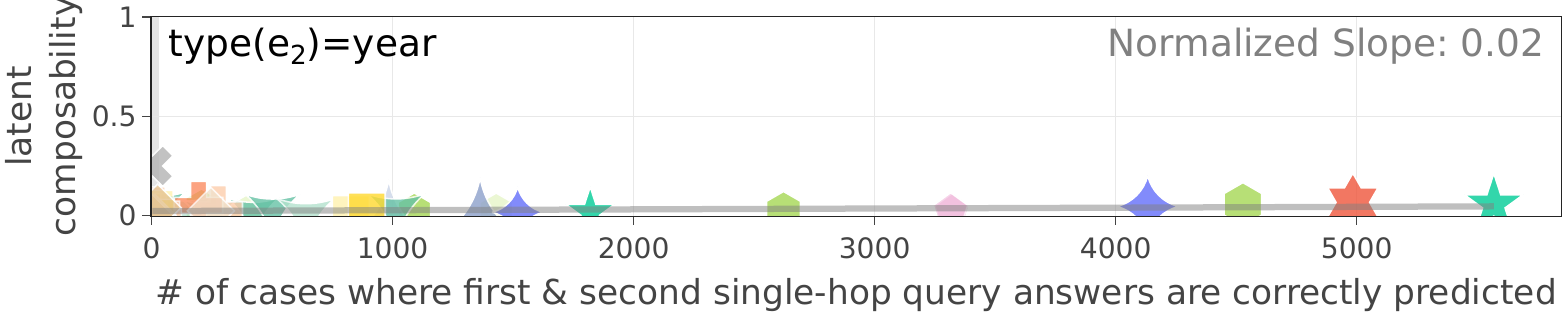}
    \end{minipage}
    \caption{Latent composability on country-type \textbf{(left)} and year-type \textbf{(right)} bridge entity subsets. Latent composability varies according to the bridge entity type; it is over 80\% for the best models when the bridge entity is a country, but around 5\% when it is a year.}
    \label{fig:latent_composability_e2}
  \end{subfigure}
  \begin{subfigure}[b]{\textwidth}
    \begin{minipage}[b]{.49\textwidth}
      \includegraphics[width=\textwidth]{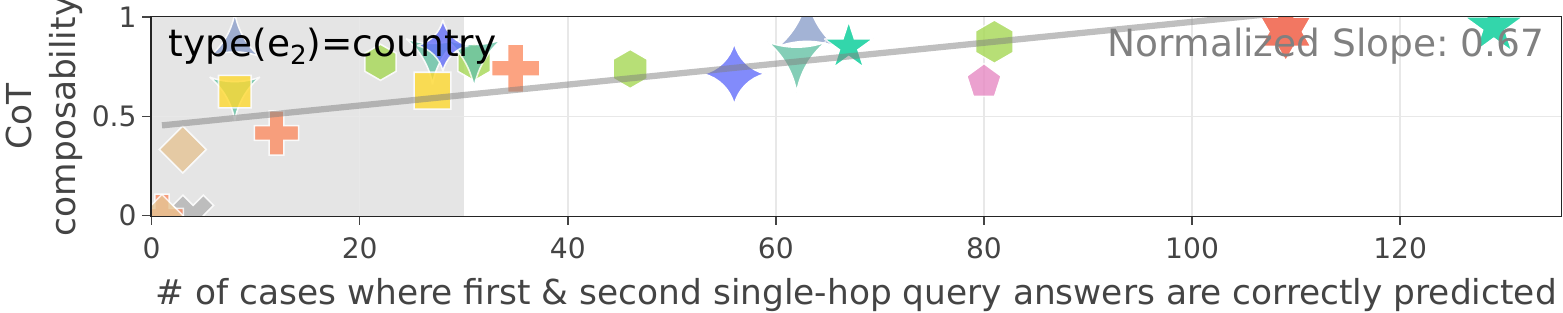}
    \end{minipage}\hfill
    \begin{minipage}[b]{.49\textwidth}
      \includegraphics[width=\textwidth]{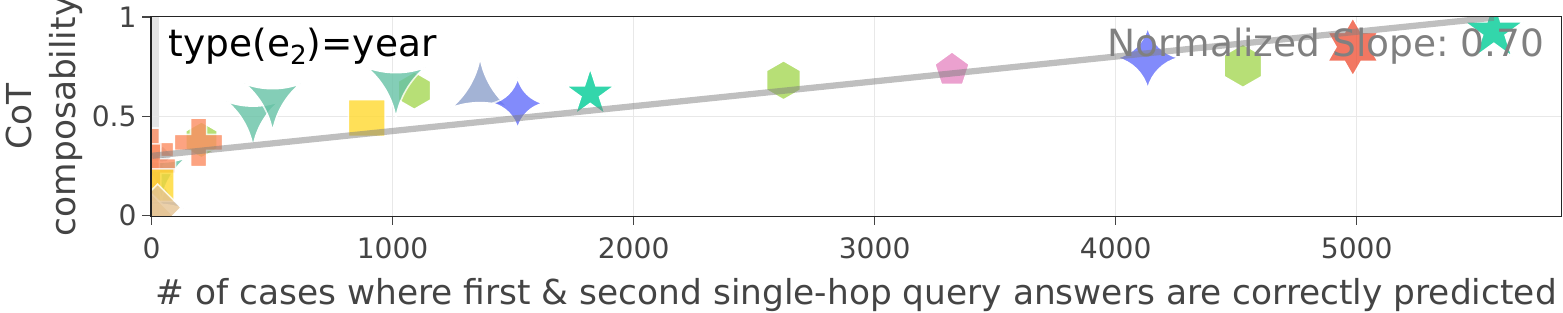}
    \end{minipage}
    \caption{CoT composability on country-type \textbf{(left)} and year-type \textbf{(right)} bridge entity subsets. CoT composability does not fluctuate as dramatically as latent composability according to the type of bridge entity.}
    \label{fig:cot_composability_e2}
  \end{subfigure}
\caption{Latent \textbf{(upper row)} and CoT \textbf{(lower row)} composability measured for subsets of the test queries in \dataset{}, grouped according to the type of the bridge entity.}
\label{fig:composability_e2}
\end{figure*}

\subsection{Latent Composability}\label{subsec:latent_composability}
We assess \emph{latent composability} as \emph{the ability of the LLM to latently compose the already-learned single-hop facts} by calculating the ratio of the cases where the LLM correctly answers the multi-hop query while correctly answering both of the corresponding single-hop queries, excluding the \guessable{} and \unusable{} cases.To check the correctness of model outputs, we use a standard Exact Match with string normalization (details in \S\ref{sec:em}).

When comparing latent composability between different models, it is misleading to compare latent composability calculated using different subsets of queries. Therefore, we calculate the ratio using the same test case subset where all of the compared LLMs correctly answer both single-hop queries where the test cases are \guessable{} or \unusable{} for neither of the models. We call this value \emph{comparative} latent composability.

\section{Experiments}

We use \dataset{} to evaluate the latent multi-hop reasoning ability of LLMs. Our results show that models can perform latent reasoning without exploiting shortcuts, but their ability depends on the type of bridge entity connecting the two facts. Table~\ref{tab:examples} in the Appendix shows exemplifying test cases, model predictions, and results.

\subsection{Experimental Setting}

We assess \nmodels{} LLMs of different families and sizes. Among the proprietary LLMs, we evaluate Claude 3.5 Sonnet~\citep{anthropic2024claude}, GPT-4o~\citep{openai2024gpt4o}, GPT-4o mini~\citep{openai2024gpt4omini}, and Gemini 1.5 Pro and Flash~\citep{Gemini-Team2024-df}. Among the open-source LLMs, we evaluate pretrained and/or instruction-tuned models of 2B to 123B parameters from the model families of Mistral~\citep{Jiang2023-lr}, Mixtral~\citep{Jiang2024-rs}, Qwen 2.5 and 2~\citep{qwen252024,Yang2024-kh}, Yi 1.5~\citep{01AI2024-tz}, Gemma 1 and 2~\citep{Mesnard2024-dk, Gemma-Team2024-fh}, and OLMo~\citep{Groeneveld2024-lc}. Refer to \S\ref{sec:inference_details} for model and inference details.

\subsection{Evaluation Results on \dataset{}}

\paragraph{There are successful cases of latent multi-hop reasoning, although the overall percentage is low.} Figure~\ref{fig:latent_composability} shows the latent composability of \nmodels{} models on \dataset{}.
While there is a meaningful number of successful latent multi-hop reasoning cases (434 for Claude 3.5 Sonnet and 438 for GPT-4o, the best performing models), the percentage of such cases of the whole dataset is low; latent composability of Claude 3.5 Sonnet and GPT-4o is only 8.4\% and 7.6\%, respectively.

\paragraph{Model scaling marginally improves overall performance.} Figure~\ref{fig:comparative_latent_composability} shows \emph{comparative} latent composability among models with different numbers of parameters within the same model family\footnote{GPT-4o vs. GPT-4o mini, Gemini 1.5 Pro vs. Gemini 1.5 Flash, Mistral Large (123B) vs. Small (22B) vs. Nemo (12B) Instruct, and Qwen 2.5 72B vs. 32B Instruct}.
We observe a consistent trend across all model families, where larger models answer more multi-hop queries correctly than smaller models, although the difference is not large in number. The gap in latent composability is 6.7\% (118) and 2.4\% (31) for GPT-4o vs. GPT-4o mini and Gemini 1.5 Pro vs. Gemini 1.5 Flash, respectively.

\paragraph{Latent composability performance varies across bridge entity types.}
Figure~\ref{fig:latent_composability_e2} shows the latent composability for two out of the four bridge entity types where the size of the subset of the queries is statistically significant (the results for all four types are in Appendix Figure~\ref{fig:latent_composability_e2_full}). Notably, the latent composability of Claude 3.5 Sonnet and GPT-4o reaches 82.6\% and 84.5\%, respectively, when the single-hop facts are connected with country-type bridge entities, but only 6.7\% and 5.7\% when the facts are connected with year-type bridge entities. The rate of improvement in latent composability with the number of known single-hop facts also varies across different bridge entity types. Our finding implies that it is important to consider the dataset distribution and perform per-relation-composition analysis when evaluating latent multi-hop reasoning (explanation in \S\ref{sec:distribution_importance}). Drawing from prior works, we speculate that the high composability of country-related queries might stem from more frequent exposure to learning country-related facts together or in composition during pretraining (explanation in \S\ref{sec:why_bridge}).

\paragraph{There exist significant disparities between CoT and latent composability.} While GPT-4o achieves 92.8\% composability with CoT reasoning, it is only 7.6\% with latent reasoning (Figure~\ref{fig:cot_composability}, \ref{fig:latent_composability}), with almost no cases where latent reasoning succeeds but CoT fails (Appendix Figure~\ref{fig:composability_ratio}). Models that know more single-hop facts and larger models show dramatic improvements in composability for CoT reasoning compared to latent reasoning (Figures~\ref{fig:comparative_cot_composability}, \ref{fig:comparative_latent_composability}), suggesting that merely increasing parameter count cannot effectively enhance latent multi-hop reasoning. Furthermore, CoT composability remains relatively consistent across bridge entity types (Figure~\ref{fig:cot_composability_e2}, \ref{fig:latent_composability_e2}).
Drawing from prior works, we speculate that the explicit generation of the bridge entity is the main factor behind high CoT composability (explanation in \S\ref{sec:why_cot}).

\subsection{Additional Analysis}\label{sec:additional-analysis}

\begin{figure}[t!]
  \centering
  \includegraphics[width=0.48\textwidth]{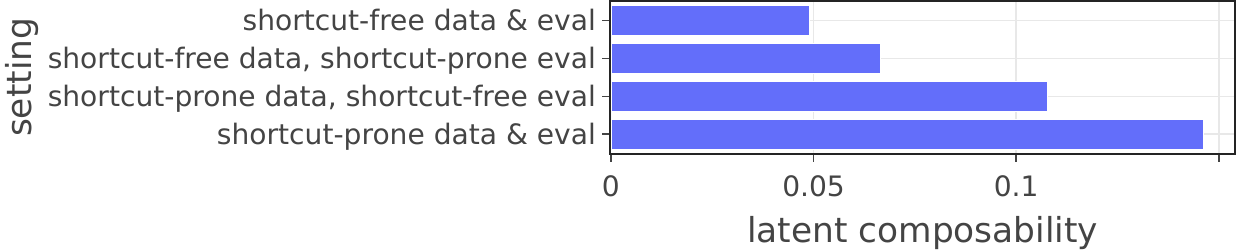}
\caption{Latent composability measured with shortcut free/prone data and evaluation, averaged across models shown in Appendix Figure~\ref{fig:shortcut_free_vs_prone}. Latent composability measured with \dataset{} and the proposed evaluation procedure is about three times lower than the shortcut-prone counterpart.}
\label{fig:shortcut_free_vs_prone_summary}
\end{figure}

\paragraph{Shortcut-free evaluation is important.}
To check the importance of addressing shortcuts, we perform a comparative experiment with a shortcut-prone dataset and evaluation procedure that does not follow the proposed desiderata. 
Specifically, we construct a dataset with almost exactly the same distribution of the relation composition types (and thus the bridge entity types) of \dataset{}, but without applying any measure to remove potential shortcuts such as the entity co-occurrence-based filtering or relation-specific heuristics. As the evaluation procedure, we only check whether both of the single-hop facts are known by the model, and do not exclude the \guessable{} and \unusable{} cases.

Latent composability measured with shortcut-free data and evaluation procedure is three times lower than the shortcut-prone counterpart (Figure~\ref{fig:shortcut_free_vs_prone_summary}), and the former is consistently lower than the latter across all models (Appendix Figure~\ref{fig:shortcut_free_vs_prone}). These results imply that overlooking shortcut exploitation can overestimate the latent composability of the model.

\begin{table}[t!]
\centering
\resizebox{0.48\textwidth}{!}{
\begin{tabular}{l@{\hspace{-2em}}rrrrr}
\toprule
\textbf{Relation} & \textbf{Shortcut-prone} & \textbf{Shortcut-prone} & \textbf{Shortcut-free} & \textbf{Total} \\
\textbf{Composition Type} & \textbf{Ratio (\%)} & \textbf{Count} & \textbf{Count} & \textbf{Count} \\
\midrule
person-country-callingcode & 97.44 & 456 & 12 & 468 \\
person-country-capital & 100.00 & 468 & 0 & 468 \\
person-country-ccn3 & 94.23 & 441 & 27 & 468 \\
person-country-currency & 88.46 & 414 & 54 & 468 \\
person-country-currencyshort & 93.38 & 437 & 31 & 468 \\
person-country-currencysymbol & 76.28 & 357 & 111 & 468 \\
person-country-estcommonname & 64.32 & 301 & 167 & 468 \\
person-country-jpncommonname & 71.58 & 335 & 133 & 468 \\
person-country-roundedlat & 100.00 & 468 & 0 & 468 \\
person-country-roundedlng & 100.00 & 468 & 0 & 468 \\
person-country-ruscommonname & 75.00 & 351 & 117 & 468 \\
person-country-spacommonname & 97.22 & 455 & 13 & 468 \\
person-country-tld & 98.72 & 462 & 6 & 468 \\
person-country-urdcommonname & 20.51 & 96 & 372 & 468 \\
person-year-masterschampion & 81.24 & 589 & 136 & 725 \\
person-year-nobellit & 87.57 & 620 & 88 & 708 \\
person-year-uspresident & 98.45 & 697 & 11 & 708 \\
\midrule
{total} & {85.30} & {7,415} & {1,728} & {8,693} \\
\bottomrule
\end{tabular}}
\caption{Subject-object co-occurrence statistics of Compositional Celebrities (CC)~\citep{Ofir-Press2023-dm} dataset in Dolma v1.5 using the WIMBD API.}
\label{tab:previous-dataset}
\end{table}

\paragraph{Explicit grounding to the co-occurrence statistics is important.}

To emphasize the need for an explicit grounding to the co-occurrence statistics in the dataset construction process, we check the co-occurrence counts of the subjects and objects in the test queries of the Compositional Celebrities~\citep{Ofir-Press2023-dm} dataset in Dolma v1.5 using the WIMBD (What's In My Big Data)~\citep{Elazar2024-ct} API. Table~\ref{tab:previous-dataset} shows that 85.30\% of the queries in the dataset are prone to subject-object shortcuts, appearing at least once in Dolma v1.5. More than 95\% of the data subsets are prone to subject-object shortcuts for 7 out of 17 categories, where three categories are completely unusable with a ratio of 100\%. Note that we have not even used all possible aliases of the subjects and objects, and that we have used only Dolma v1.5 to measure the co-occurrence counts; the percentage can rise even further when all aliases and more pretraining datasets are considered. Considering the relation-object shortcuts will make even more data examples unusable for shortcut-free evaluation of latent multi-hop reasoning.\looseness-1

\begin{figure*}[t!]
  \centering
  \begin{subfigure}[b]{.95\textwidth}
  \includegraphics[width=\textwidth]{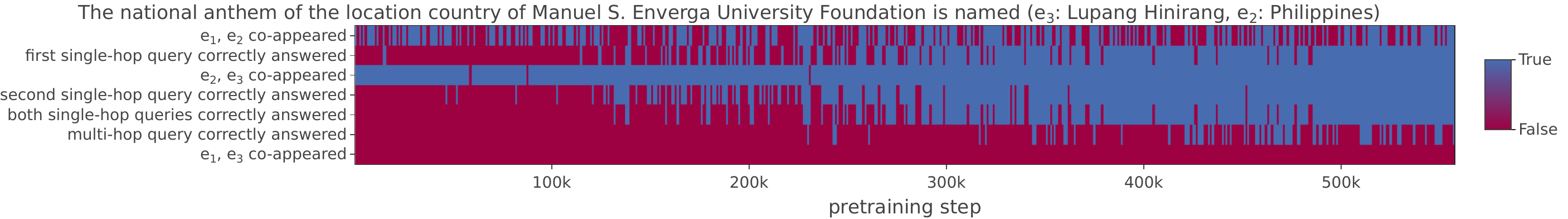}
  \end{subfigure}
\caption{One successful case of OLMo 7B that demonstrates the emergence of latent multi-hop reasoning ability even when \ea{} and \ec{} have never co-appeared in any sequence throughout pretraining.
While being pretrained for 557K steps, the model starts to correctly predict the answer to the single-hop queries after consistently seeing (\ea{}, \eb{}) and (\eb{}, \ec{}) together across multiple pretraining steps. After the model starts to learn to correctly answer the single-hop queries, the model starts to learn to correctly answer the multi-hop query.}
\label{fig:olmo_success}
\end{figure*}

\begin{figure}[t!]
  \centering
  \begin{subfigure}[b]{.22\textwidth}
  \includegraphics[width=\textwidth]{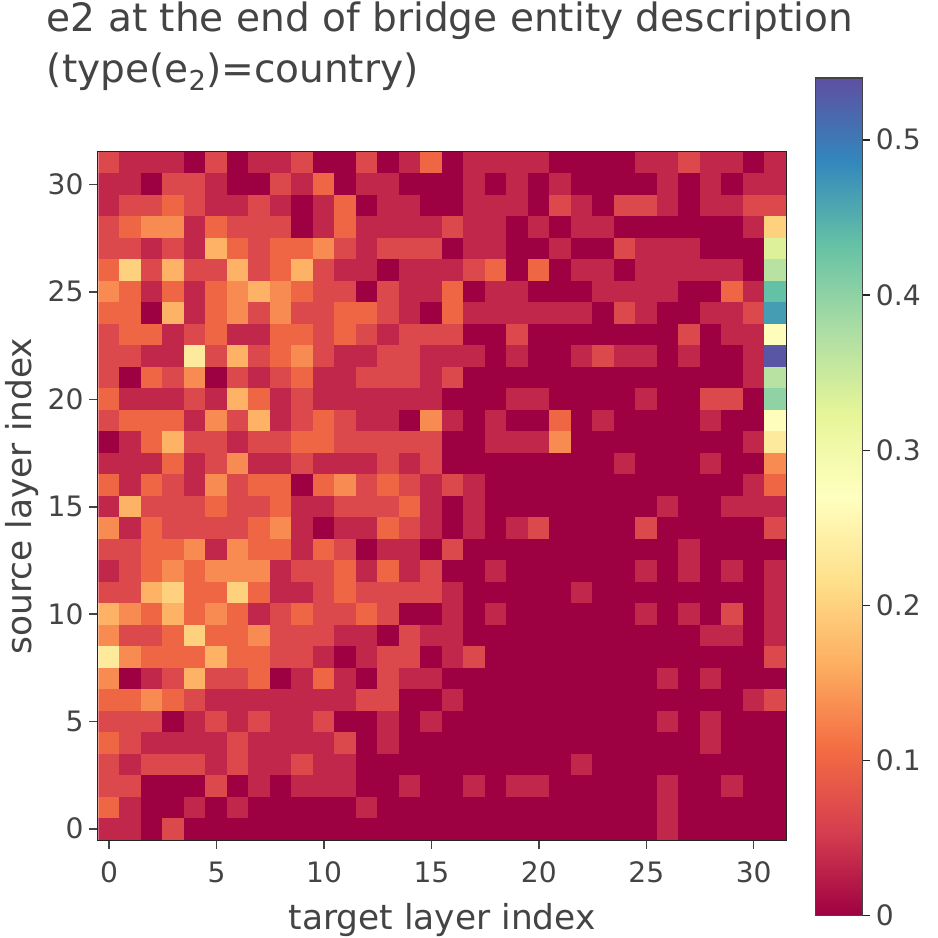}
  \end{subfigure}\hfill
  \begin{subfigure}[b]{.22\textwidth}
  \includegraphics[width=\textwidth]{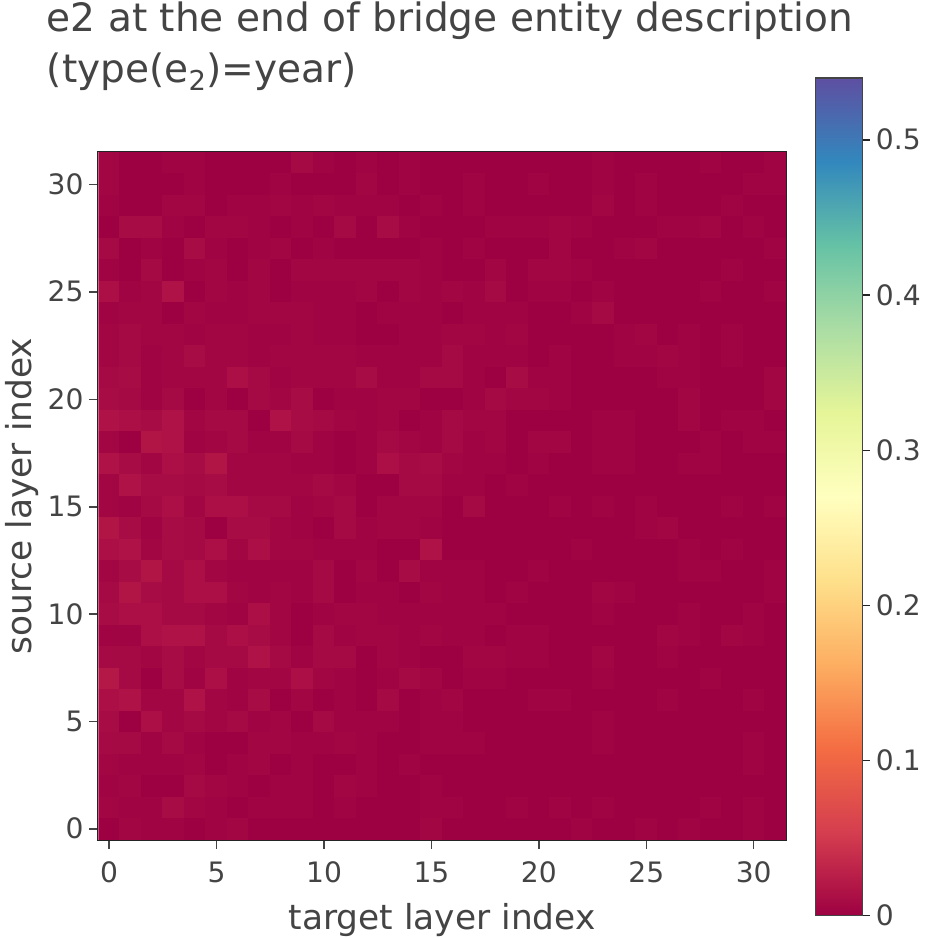}
  \end{subfigure}
\caption{Experimental results with Mistral 7B v0.3 using Patchscopes~\citep{Ghandeharioun2024-lm} to examine whether the model constructs latent representations of the bridge entity (e.g., \textify{1984} for \textify{the year Scarlett Johansson was born}) during the multi-hop query processing. Latent representations of bridge entities are constructed more often for queries with country-type bridge entities (that have higher latent composability).}
\label{fig:patchscopes_summary}
\end{figure}

\paragraph{Latent representation of the bridge entity appears more often for query types with higher latent composability.}
We perform an experiment using Patchscopes~\citep{Ghandeharioun2024-lm} following \citet{Biran2024-gw} using Mistral 7B v0.3, which examines whether the model constructs latent representations of the bridge entity (e.g., \textify{1984}) for its descriptive mention (\textify{the year Scarlett Johansson was born}) encountered during multi-hop query processing.
Figure~\ref{fig:patchscopes_summary} shows how often the hidden states at the last token of the descriptive mention taken from different layers (y-axis) generate the bridge entity when patched into appropriate contexts at different layers (x-axis), for the queries with country/year-type bridge entities where both single hop facts are known by the model. The bridge entity is generated more often (which suggests that the latent bridge entity representations are constructed more often) for the queries with higher latent composability (queries with country-type bridge entities). Details are in \S\ref{sec:patchscopes}.\looseness-1

\paragraph{Emergence of latent multi-hop reasoning during pretraining.}
Our analysis of OLMo 7B's intermediate checkpoints (557 checkpoints from 1K to 557K pretraining steps) shows that for a subset of prompts, OLMO first learns to predict the single-hop answers, and then begins to correctly answer the respective multi-hop query. Figure~\ref{fig:olmo_success} illustrates one such case. This set is small: 12 out of 13 cases where the model successfully performs multi-hop reasoning at a point, among 110 cases where the model is correct on both single-hop facts at some point and the model is not likely to be guessing the answer at any point during pretraining. That said, for OLMo, we have access to all the pretraining sequences in order and can hence guarantee that head entities have not been seen together with the answer entities in any of the pretraining sequences. Together with our other filtering to reduce the probability that answers can be correct by chance and guesswork, this indicates that even a small model like OLMo 7B can perform some level of latent reasoning, albeit only occasionally. More details of the experiment are provided in \S\ref{sec:olmo}.

\section{Conclusion}

We outline desiderata for shortcut-free evaluation of LLMs' ability to latently recall and compose learned single-hop facts to answer multi-hop queries. By filtering entity co-occurrences and systematic removal of potential shortcuts, we construct the \dataset{} dataset, enabling a rigorous assessment of latent multi-hop reasoning. Our analysis reveals that while models can perform latent reasoning effectively in specific scenarios, their ability varies dramatically across different types of queries. This fluctuation, along with the significant gap between latent and explicit CoT reasoning, suggests substantial room for improvement in how LLMs internally compose their knowledge. Our work provides resources and insights for precise evaluation, understanding, and improvement of latent multi-hop reasoning of LLMs.

\section*{Limitations}
We do not test other forms of compositional reasoning such as comparisons because if the answer is binary, it is hard to rule out the cases of guessing.
We do not test more than two hops because latent two-hop composability is already quite low, and adding more complexity to the problem may lower it to zero success cases.
We do not consider facts that are subject to frequent change over time to compare models trained at different corpus cutout times.
While we cannot know whether the LLMs we evaluate have not been trained on synthetic data generated from a knowledge graph, the low latent composability and high CoT composability obtained with our dataset suggest that the chance is very low for the evaluated models.
While we cannot guarantee that the head and answer entities of every test query of \dataset{} would have never been learned in a single pretraining sequence for every model that we evaluate, we believe that our dataset construction that utilizes document co-occurrence counts of multiple pretraining corpora provides a tight approximation, which is also supported by the experimental results in \S\ref{sec:search}.

\section*{Acknowledgments}
We would like to thank Sang-Woo Lee, Hoyeon Chang, and Chris Dyer for the valuable feedback and discussions. We are grateful to Yanai Elazar for their support of the WIMBD~\citep{Elazar2024-ct} API throughout the project.

\bibliography{acl_latex}

\appendix

\begin{table*}[ht!]
\centering
\resizebox{0.8\textwidth}{!}{\begin{tabular}{ll}
\toprule
\textbf{Relation Composition Type} & \textbf{Heuristic Conditions to Meet} \\ \midrule
person-birthcity-year                          & city $\neq$ capital of a country                                             \\
                                          & person's birth year $\neq$ event year                                        \\
person-undergraduniversity-year                    & person's birth year $\neq$ university's inception year                       \\
person-birthyear-winner                        & person's birth/citizenship country $\neq$ winner's birth/citizenship country \\
person-birthyear-event country/city/leader     & person's birth/citizenship country $\neq$ event country                      \\
university-inceptionyear-winner                    & university's location country $\neq$ winner birth/citizenship country        \\
                                          & university $\neq$ winner's university for any degree                         \\
city-eventyear-winner                     & event country $\neq$ winner's birth/citizenship country                      \\
university-inceptionyear-event country/city/leader & university's location country $\neq$ event country                           \\ \bottomrule
\end{tabular}}
\caption{Relation-specific heuristic conditions that the collected test cases need to satisfy to prevent using the test queries where spurious correlations of \ea{} and \ec{} exist.}
\label{tab:relation_heuristics}
\end{table*}

\section{Details of Dataset Construction}

\subsection{Implementation of steps 1-2}\label{sec:impl12}

We choose Wikidata~\citep{Vrandecic2014-eu} as the knowledge graph and collect facts in English. To ensure fair evaluation across models with different knowledge cutoff dates, we abstain from using relations subject to change over time (\relationify{organization-CEO} or \relationify{person-spouse}). We select 11 types of $\fa{} = \eb$ and 10 types of $\fb{} = \ec$ that are connected to each other with four types of \eb{} -- \textify{country}, \textify{city}, \textify{university}, and \textify{year} -- resulting in a total of 21 relation composition types, although the number reduces to 17 through filtering during the implementation of step 3. Table~\ref{tab:dataset_stats_subtype} shows the 17 relation compositions that consist \dataset{}. As shown in the table, the relation compositions are divided further into subtypes for events\footnote{Olympics Summer, Olympics Winter, Eurovision, Champions League Final, G7, European Capital of Culture} and award winners\footnote{Nobel Prize for Peace, Chemistry, Literature, Physiology or Medicine, and Masters Championship}.

To use only the entities where \ea{} and \ec{} are not likely to be connected through other popular single-hop relations, we apply a relation-specific heuristic filtering. For instance, for \relationify{person-birthyear-eventcountry} relation (e.g., \textify{The country where the Eurovision Song Contest took place in the birth year of \ea{} is}), we exclude the cases where the country in which the event took place is the same with the person's birth country. A list of such heuristics is provided in Appendix Table~\ref{tab:relation_heuristics}.

Additionally, we exclude cases where \eb{} can be easily inferred from the surface form of \ea{}, such as \relationify{university-locationcountry(University of Washington) = United States}, as these cases are where the multi-hop query is reduced into more like a single-hop query and also the cases where a substring of \ea{} has relatively higher chance to frequently co-appear with \ec{} in the same training sequences. A more detailed explanation is provided in \S\ref{sec:shortcut}. Through this process, we collect about 100K tuples $(\ea, \eb, \ec, \raa, \rbb, \eea, \eeb, \eec)$, where \ea{}, \eb{}, \ec{} are the Wikidata entity titles, and \eea{}, \eeb{}, \eec{} are their corresponding sets of Wikidata aliases.

We manually construct four natural language templates for each subtype of \rbb{} and \raa{} and randomly select one of the templates for each test case. Since multi-hop queries are constructed with the combinations of the templates chosen for \rbb{} and \rbb{}, 16 templates are used for each type of relation composition. We feed \eb{} to a template for \rbb{} to create \qb{}, and feed the descriptive mention $\mu$ of the bridge entity (e.g., \textify{the year Scarlett Johansson was born}) to create \qc{}, that both take \eec{} as the answer set (e.g., $\{$\textify{United States}, \textify{US}, $\cdots \}$). The descriptive mention of the bridge entity is created with a template for \raa{}. The same descriptive mention $\mu{}$ is also used to create \qa{} in the form \textify{$\mu$ is} which takes \eeb{} as the answer set (e.g., $\{$\textify{1948}$\}$). We filter out the dataset further during quality assurance (\S\ref{sec:dataset_qa}).

\subsection{Implementation of step 3 with document co-occurrence counts}\label{sec:impl3}

Step 3 can be directly implemented when one has access to the LLM's pretraining sequences. However, we cannot use the approach as-is since LLMs are trained on different pretraining data, and this data is often not publicly available. Moreover, even if the data is available, information on how it has been broken into training sequences is rarely provided. 
To overcome these challenges, instead of computing the exact co-occurrence counts, we approximate them using two simplifications.

The first simplification is that we check the co-occurrence of the aliases of \ea{} and \ec{} \textit{within a document instead of a training sequence}. 
To be specific, we use only test cases where there is no pretraining document in which any of the possible combinations of the aliases of \ea{} and \ec{} appear together. Filtering out test cases with non-zero document co-occurrence count imposes a stricter condition than doing so with sequence co-occurrence count because training sequences are substrings of a document in most LLMs trained with document boundaries, which is the standard approach in pretraining LLMs~\citep{Zhao2024-tt}.

The second simplification is that we utilize a \textit{proxy corpus} to check the document co-occurrence since even the document-level information of the pretraining data of most models is not available.We use six different training corpora: Dolma v1.5, v1.7~\citep{Soldaini2024-zr}, Tulu v2~\citep{tulu2023ai2}, OSCAR~\citep{Suarez2020-jn}, C4~\citep{Raffel2020-nd}, and OpenWebText~\citep{openwebtext2019}, each of which contains 4,367M, 2,532M, 326K, 432M, 365M, 8M documents, used to train OLMo, OLMo 0724, OLMo Instruct~\citep{Groeneveld2024-lc}, BLOOM~\citep{BigScience-Workshop2022-ye}, T5~\citep{Raffel2020-nd}, and GPT-2~\citep{radford2019language}, respectively. The number of unique documents from the proxy corpus is roughly 4.8B.\footnote{4,367M (Dolma v1.5) + 432M (OSCAR) + 8M (OpenWebText) - 10M (OSCAR-Dolma overlap from \href{https://wimbd.apps.allenai.org/}{WIMBD~\citep{Elazar2024-ct} Demo Page})} In other words, we only use the test cases where none of any possible combination of the aliases of \ea{} and \ec{} appears together in 4.8B unique documents, which imposes a highly restrictive condition that enables obtaining a tight approximation of the test queries where the head and answer entities do not co-appear in the single pretraining sequence. While we cannot guarantee the exclusion of all entity co-occurring cases without access to exact pretraining corpora, we further validate our approximation using Google Search to check for co-occurrences across the whole web (\S\ref{sec:search}).

We use the WIMBD (What's In My Big Data)~\citep{Elazar2024-ct} API to get the document co-occurrence counts of these pretraining corpora which utilize Elasticsearch~\citep{elasticsearch2010} as the backend with case-insensitive string match and exclude the test cases with non-zero co-occurrence count. After this filtering process, we obtain about 32K test cases of 17 relation composition types in total. Note that the distribution of the relation compositions is forced to be imbalanced as \ea{} and \ec{} of some relation compositions frequently appear together in the same document and most of the test cases are removed by the co-occurrence-based filtering. We down-sample the test cases with year-type bridge entities as the queries of these types outweigh other types.

\paragraph{Data statistics}
Our dataset for shortcut-free evaluation of latent multi-hop reasoning ability contains 7,232 test cases of 17 types of relation compositions connected by 4 types of bridge entities, as shown in Table~\ref{tab:dataset_stats}. Note that the distribution of relation compositions is imbalanced as \ea{} and \ec{} of some relation compositions frequently appear together in the same document and most of the test cases are removed by the co-occurrence-based filtering.

\subsection{Filtering Out the Cases with Easily Inferrable Bridge Entities}\label{sec:shortcut}
To prevent the multi-hop query from being reduced to be more similar to a single-hop query, we filter out the cases where the bridge entity is \textit{easily inferrable} from the surface form of \ea{}~\citep{Poerner2020-ok}, e.g., \relationify{university-locationcountry( University of Washington) = United States}, since Washington is a geographical location in the United States. Note that they also correspond to the cases where a \textit{substring} of \ea{} is likely to co-appear with \ec{} in the same training sequence, e.g., \relationify{university-locationcountry-anthem( University of Washington) = The Star- Spangled Banner} where \textify{Washington} likely co-occurs with \textify{The Star-Spangled Banner}.

There are two such \raa{} in our dataset: \relationify{university-locationcountry} and \relationify{person-birthcountry}. We use the prompt to GPT 3.5 turbo~\citep{openai2024chatgpt} and Claude 3 Haiku to guess the bridge entity solely from the name of the head entity for the first single-hop facts of these relation types (instruction is provided in \S\ref{sec:instruction}), and if any of these models correctly predict the answer, we exclude it from the dataset.

As a result, for the cases with country as the bridge entity type, we only use the cases where the country of location of a university or the birth country of a person is hard to guess solely from the name without knowing the correct fact, such as \relationify{university-locationcountry(The International~Graduate~School~of~English) = South Korea} and \relationify{person-birthcountry (Natalie~Portman)~=~Israel}.

\subsection{Dataset Quality Assurance}\label{sec:dataset_qa}
We apply several heuristic filterings to enhance the quality of the dataset such as excluding the cases without a natural language Wikidata title, excluding the cases with non-Unicode characters in the entity, excluding \ea{} that contain double quotation marks or slashes, removing country flag emojis from the aliases of countries, and excluding the cases where each of \ea{}, \eb{}, and \ec{} is a substring of the others. HTML characters are escaped and normalized. We discover that when all the open-source LLMs we evaluate fail to correctly answer a single-hop query, it is either because there is an error or noise in the answer set (Wikidata aliases) or the single-hop fact is not popular, and thus discard such cases from the dataset. Additionally, we use only the test cases where any alias combination of \ea{} and \eb{}, and \eb{} and \ec{} appear together in Dolma v1.5 at least once.

\begin{table*}[ht!]
\centering
\resizebox{0.99\textwidth}{!}{\begin{tabular}{llrl}
\toprule
\textbf{Relation Composition Type} & \textbf{Relation Composition Subtype} & \textbf{Count} & \textbf{Example Multi-hop Query} \\
\midrule
\multirow[c]{4}{*}{person-birthcity-eventyear} & person-birthcity-g7year & 9 & The G7 Summit was hosted in $e_1$'s birth city in the year \\
 & person-birthcity-capitalofcultureyear & 3 & The year the birth city of $e_1$ was declared as the European Capital of Culture was \\
 & person-birthcity-olympicswinteryear & 5 & The city where $e_1$ was born hosted the Winter Olympics in the year \\
 & person-birthcity-eurovisionyear & 16 & The year when the birth city of $e_1$ hosted the Eurovision Song Contest was \\
\midrule
person-birthcountry-anthem & person-birthcountry-anthem & 22 & The name of the national anthem of the birth country of $e_1$ is \\
person-birthcountry-isocode & person-birthcountry-isocode & 6 & The ISO 3166-1 numeric code of the country where $e_1$ was born is \\
university-locationcountry-anthem & university-locationcountry-anthem & 101 & The country where $e_1$ is based has the national anthem named \\
university-locationcountry-isocode & university-locationcountry-isocode & 30 & The ISO 3166-1 numeric code used for the country where $e_1$ is located is \\
university-locationcountry-year & university-locationcountry-year & 7 & The country where $e_1$ is located was established in the year \\
\midrule
person-undergraduniversity-founder & person-undergraduniversity-founder & 33 & The person who founded the university where $e_1$ studied as an undergrad is named \\
person-undergraduniversity-year & person-undergraduniversity-year & 25 & The establishment year of the university where $e_1$ studied as an undergrad is \\
\midrule
\multirow[c]{2}{*}{city-eventyear-winner} & city-eurovisionyear-nobelchem & 1 & In the year the Eurovision Song Contest took place in $e_1$, the laureate of the Nobel Prize in Chemistry was \\
 & city-g7year-nobelchem & 1 & In the year when the G7 Summit were hosted in $e_1$, the Nobel Prize in Chemistry was awarded to \\
\midrule
\multirow[c]{6}{*}{person-birthyear-eventcity} & person-birthyear-championsleaguecity & 196 & In $e_1$'s year of birth, the host city of the Champions League final was \\
 & person-birthyear-capitalofculturecity & 264 & In the year $e_1$ was born, the city that was named the European Capital of Culture was \\
 & person-birthyear-olympicswintercity & 288 & In $e_1$'s year of birth, the Winter Olympics were hosted in the city of \\
 & person-birthyear-eurovisioncity & 391 & In $e_1$'s birth year, the host city of the Eurovision Song Contest was \\
 & person-birthyear-g7city & 167 & In the year $e_1$ was born, the host city of the G7 Summit was \\
 & person-birthyear-olympicssummercity & 83 & The city where the Summer Olympics took place in $e_1$'s year of birth is \\
\midrule
\multirow[c]{5}{*}{person-birthyear-eventcountry} & person-birthyear-olympicssummercountry & 7 & In $e_1$'s birth year, the Summer Olympics were hosted in the country of \\
 & person-birthyear-championsleaguecountry & 35 & In the birth year of $e_1$, the Champions League final was hosted in the country of \\
 & person-birthyear-olympicswintercountry & 8 & In $e_1$'s birth year, the Winter Olympics were hosted in the country of \\
 & person-birthyear-g7country & 6 & The country that hosted the G7 Summit in $e_1$'s birth year is \\
 & person-birthyear-eurovisioncountry & 68 & The country where the Eurovision Song Contest took place in the birth year of $e_1$ is \\
\midrule
person-birthyear-hostleader & person-birthyear-hostleader & 260 & The person who was the host leader of the G7 Summit in $e_1$'s year of birth is \\
\midrule
\multirow[c]{6}{*}{person-birthyear-winner} & person-birthyear-nobelpsymed & 655 & In the birth year of $e_1$, the Nobel Prize in Physiology or Medicine was awarded to \\
 & person-birthyear-nobelphysics & 675 & The winner of the Nobel Prize in Physics in the year $e_1$ was born is \\
 & person-birthyear-nobelchem & 853 & In the birth year of $e_1$, the Nobel Prize in Chemistry was awarded to \\
 & person-birthyear-nobellit & 777 & In the birth year of $e_1$, the Nobel Prize in Literature was awarded to \\
 & person-birthyear-masterschampion & 931 & In the year $e_1$ was born, the winner of the Masters Tournament was \\
 & person-birthyear-nobelpeace & 593 & The Nobel Peace Prize in the year $e_1$ was born was awarded to \\
\midrule
\multirow[c]{6}{*}{university-inceptionyear-eventcity} & university-inceptionyear-championsleaguecity & 14 & In the year $e_1$ was founded, the Champions League final was hosted in the city of \\
 & university-inceptionyear-eurovisioncity & 17 & The city that hosted the Eurovision Song Contest in $e_1$'s inception year is \\
 & university-inceptionyear-olympicswintercity & 7 & The city that hosted the Winter Olympics in the inception year of $e_1$ is \\
 & university-inceptionyear-g7city & 9 & In the inception year of $e_1$, the host city of the G7 Summit was \\
 & university-inceptionyear-olympicssummercity & 2 & In the year $e_1$ was founded, the host city of the Summer Olympics was \\
 & university-inceptionyear-capitalofculturecity & 13 & The city that became the European Capital of Culture in the founding year of $e_1$ was \\
\midrule
\multirow[c]{3}{*}{university-inceptionyear-eventcountry} & university-inceptionyear-eurovisioncountry & 7 & The country that hosted the Eurovision Song Contest in the inception year of $e_1$ is \\
 & university-inceptionyear-championsleaguecountry & 5 & The country where the Champions League final took place in the inception year of $e_1$ is \\
 & university-inceptionyear-g7country & 1 & In the year $e_1$ was founded, the host country of the G7 Summit was \\
\midrule
university-inceptionyear-hostleader & university-inceptionyear-hostleader & 9 & In the year $e_1$ was founded, the host leader of the G7 Summit was \\
\midrule
\multirow[c]{5}{*}{university-inceptionyear-winner} & university-inceptionyear-nobellit & 159 & In the inception year of $e_1$, the laureate of the Nobel Prize in Literature was \\
 & university-inceptionyear-nobelchem & 156 & The winner of the Nobel Prize in Chemistry in the year $e_1$ was founded is \\
 & university-inceptionyear-nobelpeace & 109 & The Nobel Peace Prize in the inception year of $e_1$ was awarded to \\
 & university-inceptionyear-nobelphysics & 103 & The winner of the Nobel Prize in Physics in the founding year of $e_1$ is \\
 & university-inceptionyear-nobelpsymed & 105 & In the year $e_1$ was founded, the Nobel Prize in Physiology or Medicine was awarded to \\
\midrule
total & & 7,232 & \\
\bottomrule
\end{tabular}}
\caption{Dataset statistics and example multi-hop test queries. The head entities are replaced with \ea{} to prevent potential data leakage.}\label{tab:dataset_stats_subtype}
\end{table*}

\section{Details of Evaluation Procedure}

\subsection{Excluding \unusable{} Cases}\label{sec:cheat}
We observe that there are test cases such that the LLM generation is evaluated as correct, but the test cases are actually \unusable{} for correct evaluation of the latent multi-hop reasoning due to the way the model generates the answer. The first type of such case is when the model completes the query as if constructing the answer choices of a multiple-choice question, such as completing \textify{National anthem of Woodie Flowers's country of birth:} with \textify{1. ``O Canada'' 2. ``The Star-Spangled Banner'' 3. ``God Save the Queen''}.\footnote{This occurs the most often with pretrained Qwen2 models possibly due to training on the corpus where a large portion consists of multiple-choice questions and answers.} When this is the case for the completion of a single-hop or multi-hop query, we mark the case \unusable{}.

The second type of \unusable{} only applies to multi-hop queries. Especially for instruction-tuned models, even though we explicitly instruct the model to directly generate the answer without the bridge entity (\S\ref{sec:suppress_cot}), the models sometimes generate the bridge entity before generating the answer, such as completing \textify{The name of the national anthem of the country where Rishi Bankim Chandra Colleges is based is} with \textify{\enterc{}The correct answer is India.\enterc{}The national anthem of India is Jana Gana Mana.} Such cases should be excluded from the evaluation of the latent multi-hop reasoning ability. Therefore, for the multi-hop queries with the EM score of 1, we additionally check if the LLM completion contains any of $\eb \in \eeb{}$ before the earliest $\ec \in \eec$. If it is the case, we mark the case as \unusable{}.

\subsection{Normalized Exact Match Score}\label{sec:em}
To check whether an LLM has correctly predicted the answer to the given test query, we use the binary score of normalized exact match score (EM). For each single-hop query, we apply string normalization to the completion of the LLM and each of the answer candidates in the answer set, which is the alias of the entity. The normalization consists of applying lowercase, removing accents, articles, and spaces in abbreviations, and replacing punctuation marks with spaces. The EM is 1 (correct) if any of the answer candidates is included in the generation respecting the word boundaries, and 0 (incorrect) otherwise. For the completion of the multi-hop queries, the calculation of EM goes through one more step of checking if the test case is \unusable{}, as detailed below.

\section{Details of Experiments}

\subsection{Details of Experimental Setting}\label{sec:inference_details}

Among the open-source LLMs, we evaluate Mistral Large 2407 Instruct (123B), Small 2409 Instruct (22B), and all the pretrained and instruction-tuned models of Mistral Nemo 2407 (12B), Mistral 7B v0.3~\citep{Jiang2023-lr}, Mixtral 8x7B v0.1~\citep{Jiang2024-rs}, Qwen 2.5 (7B, 14B, 32B, 72B)~\citep{qwen252024}, Qwen 2 (7B, 72B)~\citep{Yang2024-kh}, Yi 1.5 (6B, 9B, 34B)~\citep{01AI2024-tz} Gemma (2B, 7B)~\citep{Mesnard2024-dk}, Gemma 2 (2B, 9B)~\citep{Gemma-Team2024-fh}, and OLMo (7B)~\citep{Groeneveld2024-lc}.

For all open-source LLMs, we use vLLM~\citep{Kwon2023-ur} or HuggingFace Transformers~\citep{Wolf2020-tt} to run the inference and greedy decoding\footnote{We have also tried using random seed 0 and the decoding parameters specified in \texttt{generation\_config.json} of each model in the HuggingFace Model Hub (\href{https://hf.co}{https://hf.co}). However, the performance difference was not large; in general, pretrained models performed slightly better with greedy decoding, and instruction-tuned models performed slightly worse with greedy decoding. Therefore, we chose to evaluate the models with greedy decoding for simplification and reproducibility.} All experiments are performed with 1 to 8 40GB A100s using half precision. The proprietary LLM APIs are run with the default decoding parameters.

\begin{figure*}[t!]
  \centering
  \begin{subfigure}[b]{.49\textwidth}
  \includegraphics[width=\textwidth]{figs/latent_composability_country.pdf}
  \caption{country-type bridge entity subset}
  \label{fig:latent_composability_country}
  \end{subfigure}\hfill
  \begin{subfigure}[b]{.49\textwidth}
  \includegraphics[width=\textwidth]{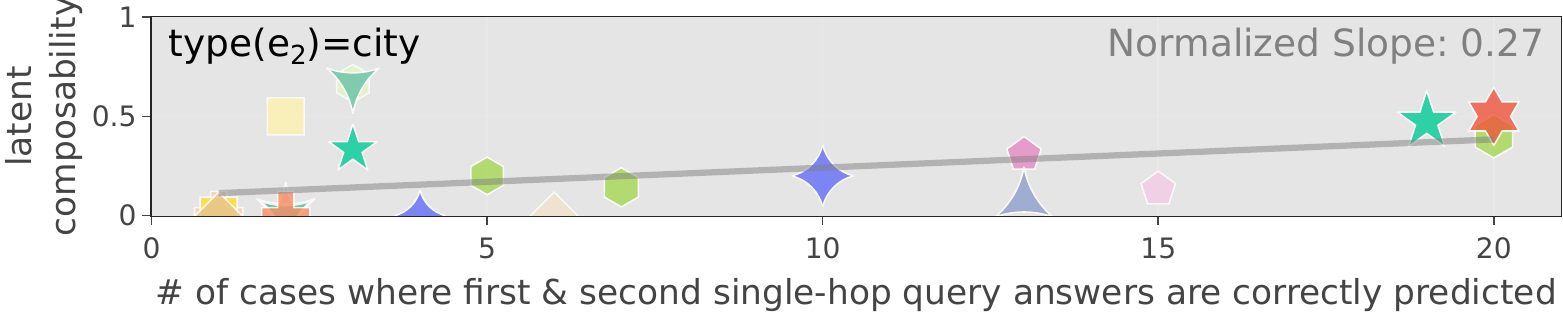}
  \caption{city-type bridge entity subset}
  \label{fig:latent_composability_city}
  \end{subfigure}
  \begin{subfigure}[b]{.49\textwidth}
  \includegraphics[width=\textwidth]{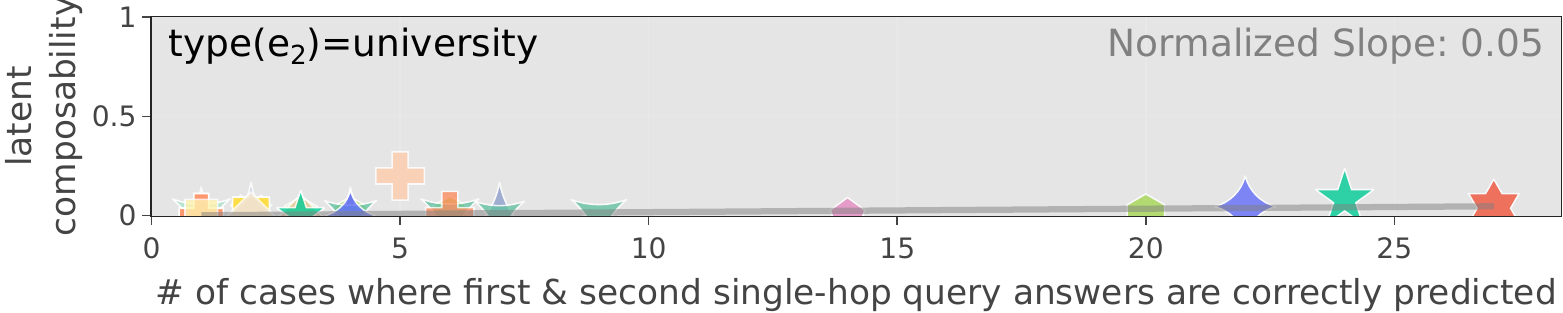}
  \caption{university-type bridge entity subset}
  \label{fig:latent_composability_university}
  \end{subfigure}\hfill
  \begin{subfigure}[b]{.49\textwidth}
  \includegraphics[width=\textwidth]{figs/latent_composability_year.pdf}
  \caption{year-type bridge entity subset}
  \label{fig:latent_composability_year}
  \end{subfigure}
\caption{Latent composability measured for subsets of the test queries in \dataset{}, grouped according to the type of the bridge entity. Latent composability varies according to the bridge entity type; it is over 80\% for the best models when the bridge entity is a country, but it is around 6\% when it is a year.}
\label{fig:latent_composability_e2_full}
\end{figure*}

\begin{figure*}[t!]
  \centering
  \begin{subfigure}[b]{.49\textwidth}
  \includegraphics[width=\textwidth]{figs/cot_composability_country.pdf}
  \caption{country-type bridge entity subset}
  \label{fig:cot_composability_country}
  \end{subfigure}\hfill
  \begin{subfigure}[b]{.49\textwidth}
  \includegraphics[width=\textwidth]{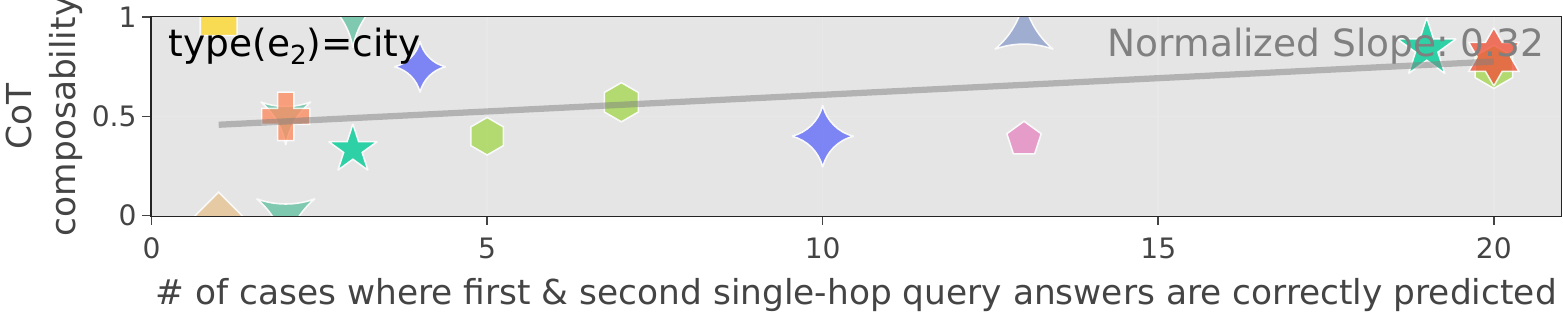}
  \caption{city-type bridge entity subset}
  \label{fig:cot_composability_city}
  \end{subfigure}
  \begin{subfigure}[b]{.49\textwidth}
  \includegraphics[width=\textwidth]{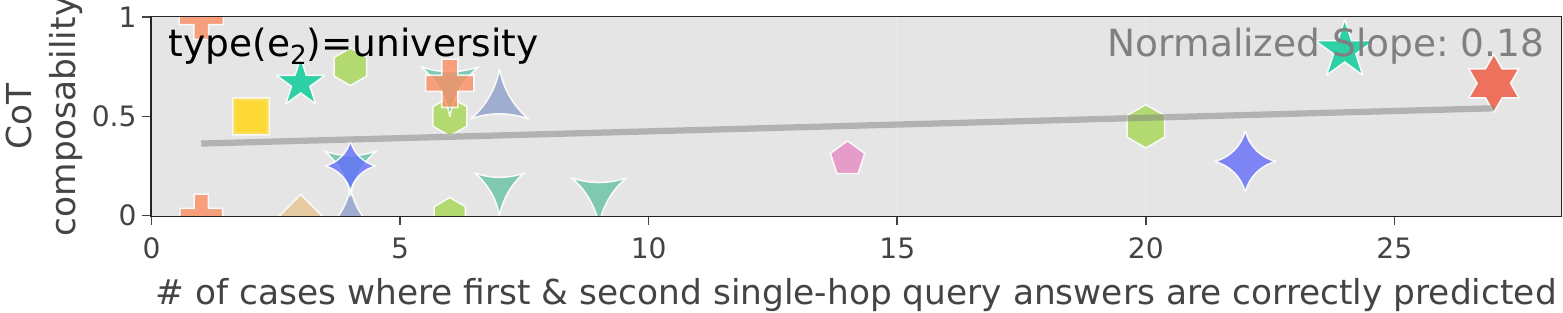}
  \caption{university-type bridge entity subset}
  \label{fig:cot_composability_university}
  \end{subfigure}\hfill
  \begin{subfigure}[b]{.49\textwidth}
  \includegraphics[width=\textwidth]{figs/cot_composability_year.pdf}
  \caption{year-type bridge entity subset}
  \label{fig:cot_composability_year}
  \end{subfigure}
\caption{CoT composability measured for subsets of the test queries in \dataset{}, grouped according to the type of the bridge entity. CoT composability does not fluctuate as dramatically as latent composability according to the type of the bridge entity, although the result is noisy for the city and university-type bridge entity subsets due to the small denominator.}
\label{fig:cot_composability_e2_full}
\end{figure*}

\begin{figure*}[t!]
  \centering
  \includegraphics[width=0.6\textwidth]{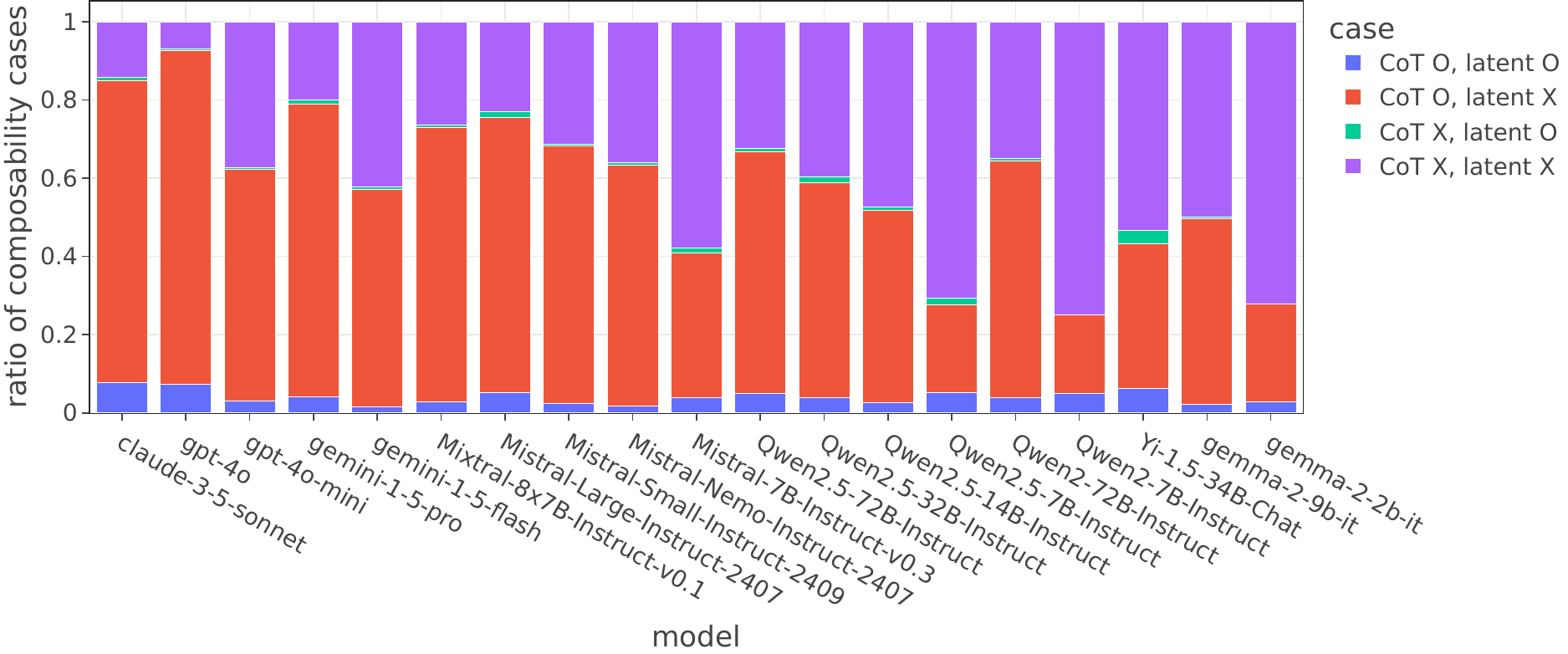}
\caption{Ratio of whether each model successfully composes the facts with latent or CoT reasoning, among the cases where the models correctly predict the answer to both single-hop questions, excluding \guessable{} and \unusable{} cases. The results are shown only when the denominator used to calculate the composability is greater or equal to 30. There are almost no cases where latent reasoning succeeds but CoT reasoning fails.}
\label{fig:composability_ratio}
\end{figure*}

\subsection{Instruction Details}\label{sec:instruction}

For the LLMs that support custom system instruction (Claude, GPT, Mistral, Qwen), we provide the instruction as the system instruction. For other models, we use the instruction at the beginning of the prompt with a separator of \textify{\enterc{}\enterc{}}.

\paragraph{CoT-suppressing instruction}
To suppress the default CoT behavior of instruction-tuned LLMs, we use the instruction \textify{Fill in the blank. Write down only what goes in the blank. Do not explain your answer. The answer can consist of multiple words.} and postpend \textify{ \_\_\_} to all test queries to measure latent composability. This prompt has the most effectively prevented the CoT-style reasoning among several different task formulations that have been manually tested.

\paragraph{CoT-triggering instruction}
To trigger CoT of instruction-tuned LLMs, we use the following instruction: \textify{Fill in the blank. First, write the step-by-step explanation necessary to get the solution with the prefix "EXPLANATION:". After that, write down the final answer with the prefix "ANSWER:". For the final answer, write down only what goes in the blank. The answer can consist of multiple words.} and postpend \textify{ \_\_\_} to the tested multi-hop query.

\paragraph{Internal think-step-by-step instruction}
We use the following instruction: \textify{Fill in the blank. Write down only what goes in the blank. Think step-by-step, but do it only internally and do not explain it in the answer. The answer can consist of multiple words.\enterc{}\enterc{}When is \ea{}'s birth year? Use the information.\enterc{}} and postpend \textify{ \_\_\_} to the tested multi-hop query.

\paragraph{Guessing the bridge entities}
For the first single-hop facts with \relationify{university-locationcountry} relations, we use the following instruction: \textify{Guessing from the name, what are the candidates of the country where ``The University of Washington'' is likely to be located? To be more specific, does ``The University of Washington'' contain the name of a location? If so, which country is the location in? Moreover, if ``The University of Washington'' contains a word of a language other than English that is used in specific countries, what are the names of those countries? Make sure to list the names of the countries guessed solely from the name.}

For the first single-hop facts with \relationify{person-birthcountry} relations, we use the following instruction: \textify{Guessing from the name, what are the candidates of the country where ``Shohei Ohtani'' was likely to be born? To be more specific, what are the candidates of the country where someone with the first name ``Shohei'' was likely to be born? Likely, what are the candidates of the country where someone with the last name ``Ohtani'' was likely to be born? Make sure to list the names of the countries guessed solely from the person name.}

\begin{figure*}[t!]
  \centering
  \includegraphics[width=0.95\textwidth]{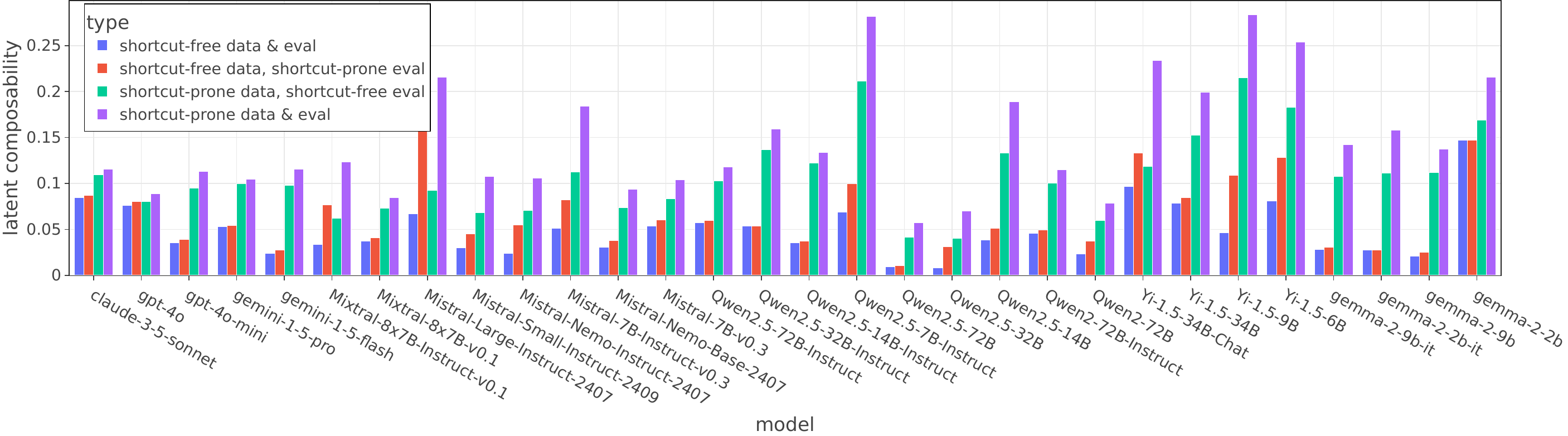}
\caption{Latent composability measured with shortcut free/prone data and evaluation. The blue bars show the latent composability on \dataset{} evaluated with the proposed evaluation procedure, while the purple bars show the latent composability on shortcut-prone evaluation. The results are shown only when the denominator used to calculate the composability is greater or equal to 30. Latent composability measured with \dataset{} and the proposed evaluation procedure is consistently lower than that measured with shortcut-prone data and evaluation across all models, implying that overlooking shortcut exploitation can lead to an overestimation of the actual latent composability.}
\label{fig:shortcut_free_vs_prone}
\end{figure*}

\subsection{Additional Google Search Filter}\label{sec:search}

For the subset of the test queries with country-type bridge entities where latent composability is notably high, we experiment with adding a Google Search filter to further exclude test cases where the head and answer entities appear together in \textit{any} of approximately 400B documents indexed by the Google Search Engine. Note that such filtering is aggressive and greatly reduces the number of test cases usable for measuring latent composability. Since the latent composability of only five models is calculated with a denominator of greater or equal to 30, we calculate the average relative change of latent composability among these five models. Denoting $c$ as the original latent composability and $c'$ as the latent composability on the subset with an additional Google Search filter, we calculate the average relative drop after applying Google Search as $\mathbb{E}{[\frac{c - c'}{c}]}$, and the value is minimal as 0.03.

\subsection{Patchscopes Experiment}\label{sec:patchscopes}

\begin{figure*}[t!]
  \centering
  \begin{subfigure}[b]{.24\textwidth}
  \includegraphics[width=\textwidth]{figs/patchscopes_mistral_t1_e2_country.pdf}
  \end{subfigure}
  \begin{subfigure}[b]{.24\textwidth}
  \includegraphics[width=\textwidth]{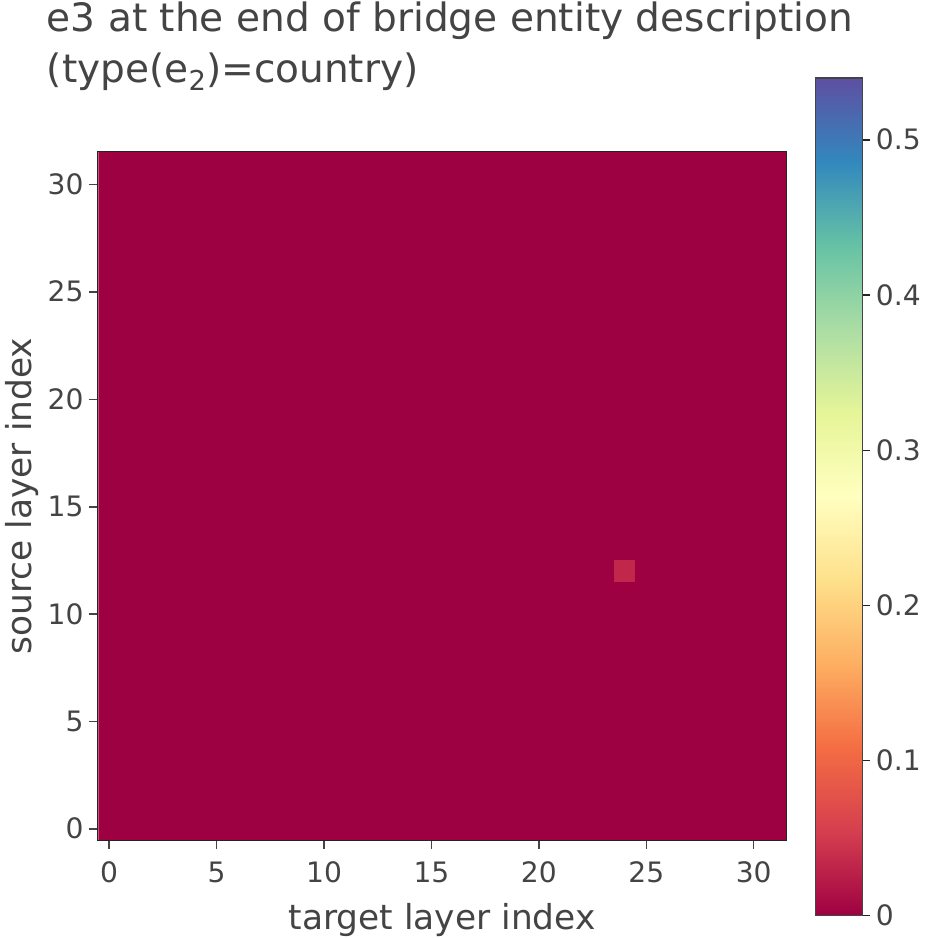}
  \end{subfigure}
  \begin{subfigure}[b]{.24\textwidth}
  \includegraphics[width=\textwidth]{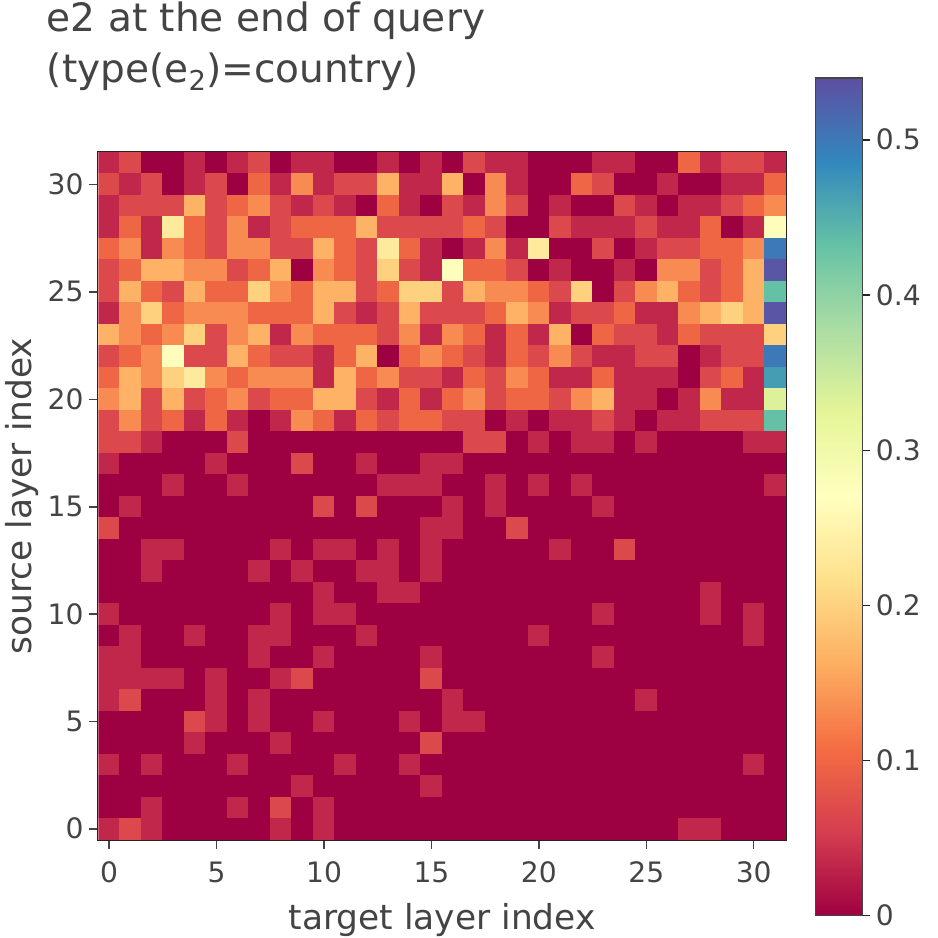}
  \end{subfigure}
  \begin{subfigure}[b]{.24\textwidth}
  \includegraphics[width=\textwidth]{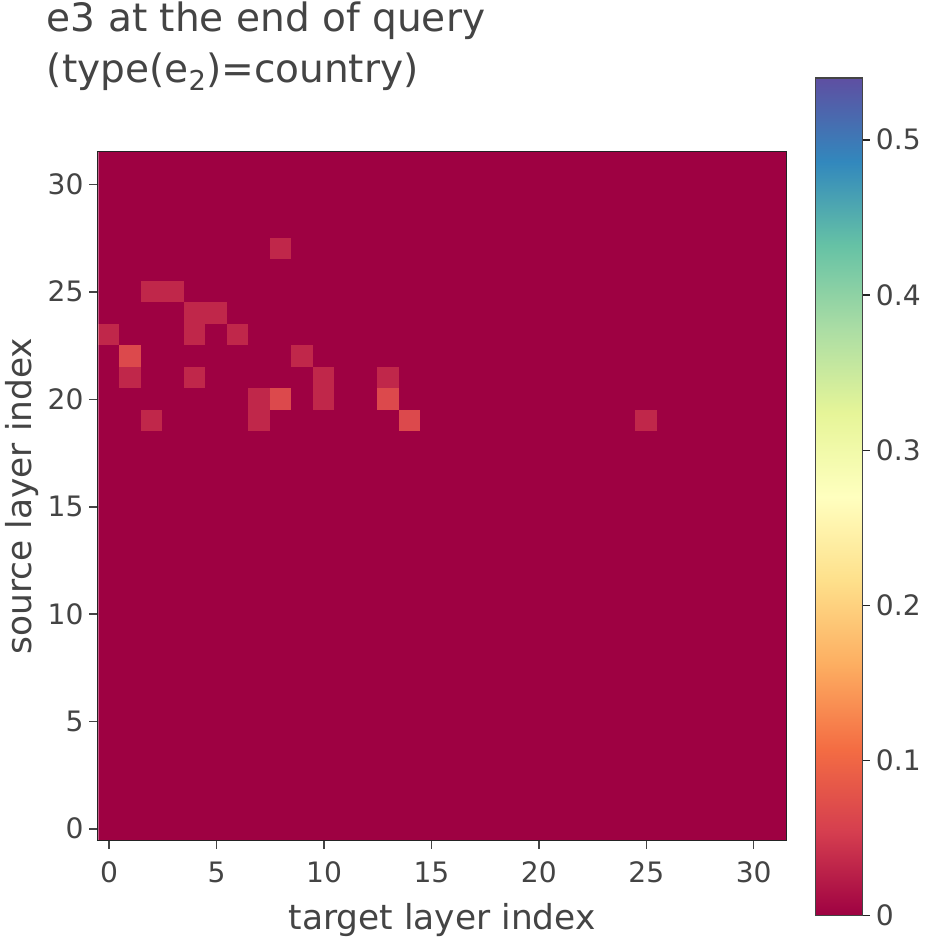}
  \end{subfigure}
  \begin{subfigure}[b]{.24\textwidth}
  \includegraphics[width=\textwidth]{figs/patchscopes_mistral_t1_e2_year.pdf}
  \end{subfigure}
  \begin{subfigure}[b]{.24\textwidth}
  \includegraphics[width=\textwidth]{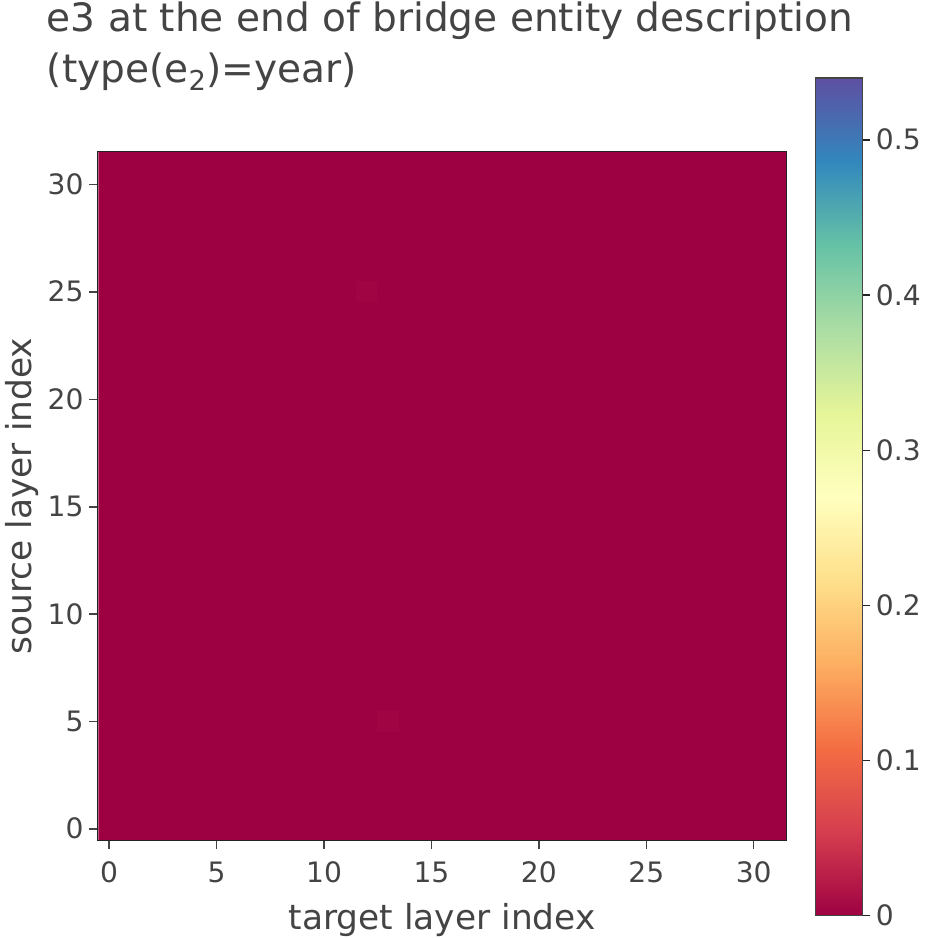}
  \end{subfigure}
  \begin{subfigure}[b]{.24\textwidth}
  \includegraphics[width=\textwidth]{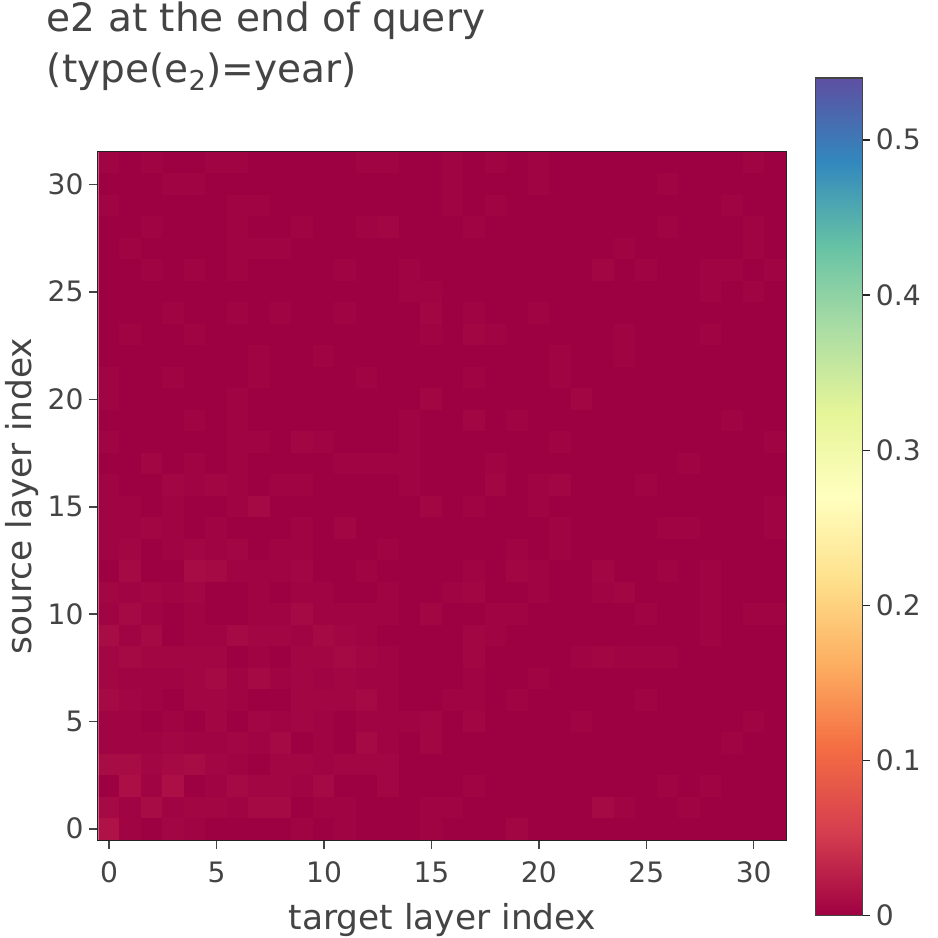}
  \end{subfigure}
  \begin{subfigure}[b]{.24\textwidth}
  \includegraphics[width=\textwidth]{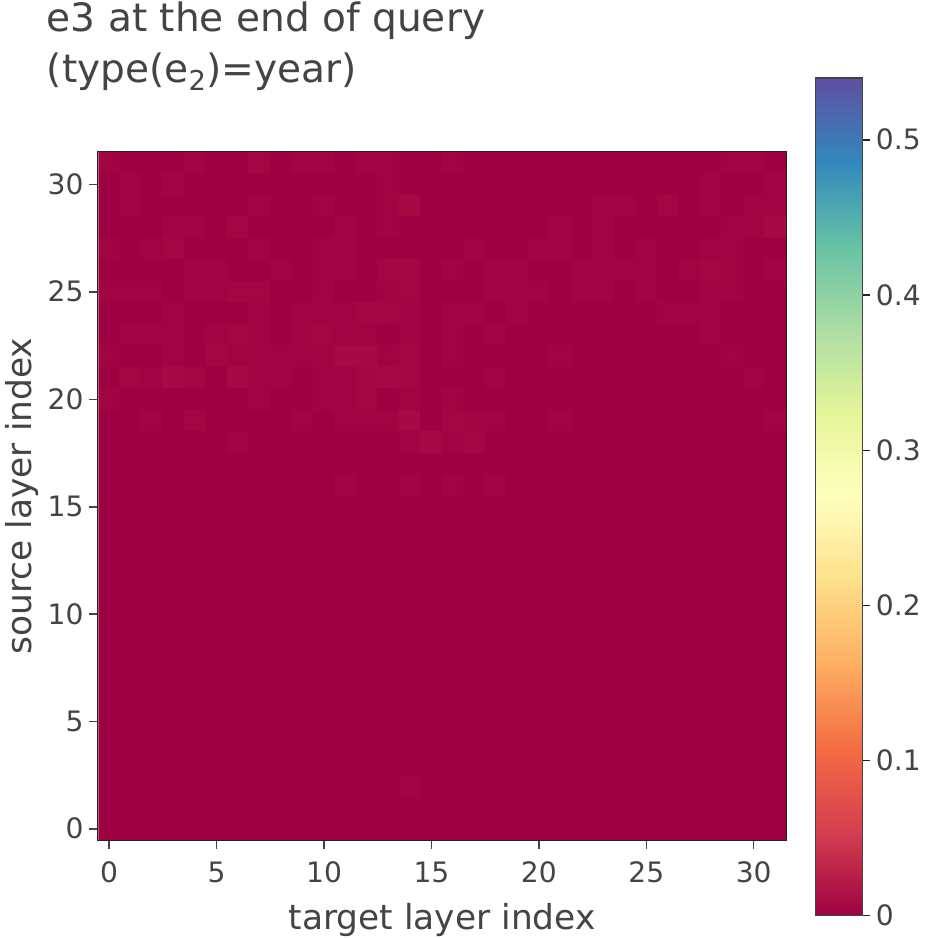}
  \end{subfigure}
\caption{Experimental results with Mistral 7B v0.3 that apply Patchscopes~\citep{Ghandeharioun2024-lm} to examine whether the model constructs latent representations of the bridge entity and answer entity at the last token of the descriptive mention of the bridge entity and the last token of the multi-hop query, for the queries with country-type bridge entities \textbf{(top)} and year-type bridge entities \textbf{(bottom)}. Latent representations of bridge entities are constructed more often for queries with country-type bridge entities (that have higher latent composability).}
\label{fig:patchscopes}
\end{figure*}

Using Patchscopes~\citep{Ghandeharioun2024-lm} following the study of \citet{Biran2024-gw}, we check how often the latent representation of the bridge entity and the answer entity is constructed at the last token of the descriptive mention of the bridge entity (e.g., \textify{the year Scarlett Johansson was born}) and the last token of the multi-hop query.

The experiment is done using the following procedure. First, we take a certain layer's hidden state computed when an LLM processes a multi-hop query (the source prompt) at either the last token position of the descriptive mention of the bridge entity or the multi-hop query. Second, we feed the target prompt \textify{StarCraft: StarCraft is a science fiction real-time strategy game, Leonardo DiCaprio: Leonardo DiCaprio is an American actor, Samsung: Samsung is a South Korean multinational corporation, x}\footnote{We slightly modify the original target prompt used in the work of \citet{Ghandeharioun2024-lm} and \citet{Biran2024-gw}, to use the description of StarCraft instead of \textify{Syria: Syria is a country in the Middle East}, in order to avoid any entity that falls into the types of the bridge entity for our dataset being used as the few-shot example and contaminating our analysis.}
into the same LLM, with activation patching~\citep{vig2020causal} of replacing a certain layer's hidden state at the token \textify{x} with the hidden state taken from the source prompt. Lastly, we let the model generate the output for the target prompt with the replaced hidden state, and check if the bridge entity or answer entity is included in the model's output. Following \citet{Biran2024-gw}, we sample three generations for each patch with a temperature of 1.0 and count it a success if the entity is included in any of the generations.

Since the target prompt follows the format of \textify{entity: entity description} where the entity description always starts by repeating the entity, if a latent representation of the entity is clearly constructed at the multi-hop query, it should be able to generate the entity from itself alone when it is patched to \textify{x} in the target prompt. We perform activation patching from each layer of the computation of the source prompt to each layer of the computation of the target layer and measure the \textit{extraction rate} of the entity; whether the output with the activation patching contains the entity. Note that the extraction is successful only when the latent representation of the entity emerges sufficiently clearly to be able to decode the entity only from the representation itself, and thus the extraction rate can be thought as a lower bound of how often the latent representation of the entity is constructed by the model while processing the multi-hop query.

Figure~\ref{fig:patchscopes} shows how often the hidden states taken from the layers at the end of the descriptive mention or the end of the source prompt (y-axis) generate the bridge or answer entity when patched into the layers of the target prompt (x-axis), for the queries with country and year-type bridge entities where both single hop facts are known by the model. The results for other types of queries are not shown due to an insufficient number of such cases. The bridge entity is generated more often, which suggests that the latent bridge entity representations are constructed more often, for the type of queries with higher latent composability (queries with country-type bridge entities).

\subsection{Emergence of Latent Multi-Hop Reasoning}\label{sec:olmo}
OLMo~\citep{Groeneveld2024-lc} provides intermediate training checkpoints (557 checkpoints from 1K to 557K pretraining steps) and the pretraining sequences that the model learns at each of the 557K steps. This allows us to: (1) definitively verify whether the head entity (\ea{}) and answer entity (\ec{}) of a multi-hop test query appear together in any single sequence during pretraining, without relying on any approximation, and (2) track the emergence of latent multi-hop reasoning ability by monitoring the model's accuracy as training progresses.

We build an ElasticSearch index using all of OLMo 7B's pretraining sequences to check whether (\ea{}, \eb{}), (\eb{}, \ec{}), and (\ea{}, \ec{}) co-appear in any single training sequence that the model learns at each pretraining step. Then, for the test queries where (\ea{}, \ec{}) never appears across all pretraining steps, we analyze the model's prediction accuracy for both single-hop and multi-hop queries. During the evaluation, we exclude any test case that is \guessable{} or \unusable{} at any of the 557 pretraining steps.

Through this evaluation procedure, we observe 13 (11.8\%) cases where the model successfully performs multi-hop reasoning at some point during pretraining, out of 110 cases where the model is correct on both single-hop facts at some point and the model is not likely to be guessing the answer at any point during pretraining. In 12 of these 13 cases, the model begins to correctly answer the multi-hop query only after learning both constituent single-hop facts. While the number of such success cases is currently limited by the model's capacity and is too small for quantitative analysis, these examples provide direct evidence that LLMs can develop latent multi-hop reasoning capabilities purely through pretraining.

Figure~\ref{fig:olmo_success} demonstrates one successful case of OLMo 7B that demonstrates the emergence of latent multi-hop reasoning ability even when \ea{} and \ec{} have never co-appeared in any sequence throughout pretraining. It is also noticeable that the model starts to correctly predict the answer to the single-hop queries after consistently seeing (\ea{}, \eb{}) and (\eb{}, \ec{}) together across multiple pretraining steps. This aligns with the finding of \citet{Chang2024-yj} that models learn simple facts by accumulating observations of the fact.

\section{Discussion}

\subsection{Importance of Considering Dataset Distribution}\label{sec:distribution_importance}
Due to our co-occurrence-based filtering process during dataset construction, year-type bridge entity cases comprise the majority of the dataset, while most test cases with other bridge entity types were filtered out. However, if the dataset had been constructed with mostly country-type bridge entity cases, the measured latent composability would have appeared much stronger. Therefore, when evaluating the latent multi-hop reasoning ability of LLMs, it is crucial to have diverse types of bridge entities, consider the distribution of the types of connected facts, and analyze performance separately for different types.

\subsection{Why Does Latent Composability Vary Across Bridge Entity Types?}\label{sec:why_bridge}

We observe that latent composability varies significantly across bridge entity types, with notably high performance when facts are connected through country-type bridge entities.

Note that it is unlikely that such a success case of latent multi-hop reasoning is obtained due to a flaw in our dataset construction process. First, our careful dataset construction, which selects only cases where countries cannot be readily inferred (\S\ref{sec:shortcut}), accounts for easy guessing of countries from entity names (e.g., \relationify{university-locationcountry(University of Washington) = United States}). Therefore, the high latent composability of the country-type bridge entity cases does not come from the model simplifying the multi-hop query into a single-hop-like problem by easily guessing the first hop. Second, it's unlikely to stem from insufficient filtering of co-occurrences: adding a Google Search filter to exclude cases where entities appear together in search results does not drop latent composability, with an average relative drop of only 0.03 (details in \S\ref{sec:search}).

Drawing from findings in finetuning studies, e.g., \citep{Jiang2022-ku,Wang2024-tu}, one speculative explanation of what has caused LLMs to develop strong composability for test queries with country-type bridge entity is that country-related facts might be more frequently learned in composition during pretraining. While these studies show that exposure to fact compositions during finetuning can improve multi-hop reasoning, we emphasize that extending these findings to pretraining remains an untested hypothesis that warrants future investigation.

\subsection{Why Is CoT Composability Much Stronger Than Latent Composability?}\label{sec:why_cot}
We conjecture that the explicit generation of the bridge entity is the main factor behind high CoT composability.
Transformer-based LLMs go through a subject enrichment process that helps models recall the attributes of the subject~\citep{Geva2023-lq,Gottesman2024-fq}.
While LLMs can develop latent representations of bridge entities (e.g., \textify{1984} from \textify{the year Scarlett Johansson was born}) in early-middle layers~\citep{Yang2024-ib, Biran2024-gw, Li2024-oy, Wang2024-tu}, these representations may appear too late or not at all~\citep{Biran2024-gw}. In contrast, when CoT reasoning generates the correct bridge entity, it ensures a clear and early contextualized representation of the bridge entity to form, facilitating retrieval of the second single-hop fact.

Supporting this hypothesis, merely instructing models to \emph{think step-by-step}~\citep{Kojima2022-jt} but \emph{only internally}, thus without generating the bridge entity, does not improve performance; Claude 3.5 Sonnet's latent composability remains low (6.1\%) even with an explicit hint to identify and utilize the information of the bridge entity (instruction shown in \S\ref{sec:instruction}). Moreover, 96.0\% of CoT failures of Claude 3.5 Sonnet stem from incorrect bridge entity generation, highlighting its crucial role.

\begin{table*}[ht!]
\centering
\resizebox{0.99\textwidth}{!}{\begin{tabular}{p{2cm}p{4cm}p{4cm}p{4cm}p{4cm}p{4cm}p{4cm}p{4cm}}
\toprule
evaluation result & \ccolor{success} & \ccolor{success} & \wcolor{failure} & \wcolor{failure} & \wcolor{\guessable{}} & \wcolor{\guessable{}} & \wcolor{\unusable{}} \\
\midrule
relation composition type & person-birthyear-winner & university-locationcountry-anthem & university-inceptionyear-winner & person-birthyear-eventcity & person-birthyear-winner & person-birthyear-winner & university-inceptionyear-winner \\
\gcmidrule{1-8}
relation composition subtype & person-birthyear-nobelchem & university-locationcountry-anthem & university-inceptionyear-nobellit & person-birthyear-eurovisioncity & person-birthyear-nobelphysics & person-birthyear-masterschampion & university-inceptionyear-nobelpsymed \\
\midrule
$E_1$ & \setify{`Ryan Cayabyab', `Raymundo Cipriano Pujante Cayabyab'} & \setify{`Royal University of Law and Economics'} & \setify{`Nyenrode Business University', `Nijenrode'} & \setify{`Chalita Suansane'} & \setify{`Turid Balke'} & \setify{`Mario Cuenca Sandoval'} & \setify{`Lagos State University', `LASU'} \\
\gcmidrule{1-8}
$E_2$ & \setify{`1954'} & \setify{`Cambodia', `Kingdom of Cambodia', `Kampuchea', `Camboya', `Campuchia', `Kamboja', `Kambodzha'} & \setify{`1946'} & \setify{`1995'} & \setify{`1921'} & \setify{`1975'} & \setify{`1983'} \\
\gcmidrule{1-8}
$E_3$ & \setify{`Linus Pauling', `Linus Carl Pauling', `Linus C. Pauling'} & \setify{`Nokor Reach'} & \setify{`Hermann Hesse', `Hermann Karl Hesse', `H. Hesse'} & \setify{`Dublin', `Baile Átha Cliath', `City of Dublin', `Dublin city', `Dublin, Ireland'} & \setify{`Albert Einstein', `Einstein', `A. Einstein'} & \setify{`Jack Nicklaus', `Golden Bear', `Jack William Nicklaus', `William Nicklaus', `Bear', `Karnak'} & \setify{`Barbara McClintock', `Eleanor McClintock'} \\
\midrule
$q(r_1(e_1))$ & Ryan Cayabyab's year of birth is & The country where Royal University of Law and Economics is based is & Nyenrode Business University's inception year is & Chalita Suansane's year of birth is & The birth year of Turid Balke is & Mario Cuenca Sandoval's birth year is & The year Lagos State University was founded is \\
\gcmidrule{1-8}
completion &  \ccolor{1954}. He is a Filipino composer, conductor, arranger, and pianist. He is also known as Mr. C. &  \ccolor{Cambodia}.

The country where Royal University of Law and Economics is based is Cambodia.

The country where Royal University of Law &  \ccolor{1946}. The university was founded by business and industry as a post-war initiative to help rebuild the country. Nyenrode is &  \ccolor{1995}. She is a Thai model and beauty pageant titleholder who was crowned Miss Universe Thailand 2016. She &  \ccolor{1921}. She was born in Norway. She is a Norwegian actress. She is best known for her role as the mother of the main &  \ccolor{1975}. He was born in Spain. He is a Spanish actor. He is best known for his role as the main character in the Spanish &  \ccolor{1983}.

The year Lagos State University was founded is 1983.

The Lagos State University (L \\
\midrule
$q(r_2(e_2))$ & In 1954, the laureate of the Nobel Prize in Chemistry was & The name of the national anthem of Cambodia is & In 1946, the Nobel Prize in Literature was awarded to & The city where the Eurovision Song Contest took place in 1995 is & In 1921, the Nobel Prize in Physics was awarded to & In 1975, the winner of the Masters Tournament was & The Nobel Prize in Physiology or Medicine in 1983 was awarded to \\
\gcmidrule{1-8}
completion &  \ccolor{Linus Pauling}. He was awarded the prize for his research into the nature of the chemical bond and its application to the elucidation of the structure &  ``\ccolor{Nokor Reach}'' which means ``Nation of Glory''. It was adopted in 1941. The lyrics were written by Ch &  \ccolor{Hermann Hesse}, a German-born Swiss novelist and poet. Hesse is best known for his novels Siddhartha, Steppenw &  the capital of the Republic of Ireland. \ccolor{Dublin} is the largest city in Ireland and is located on the east coast of the island. The city is divided into two &  \ccolor{Albert Einstein} for his discovery of the law of the photoelectric effect. The photoelectric effect is the emission of electrons from a metal surface when light &  \ccolor{Jack Nicklaus}. He won the tournament for the fifth time, and he did it in style. Nicklaus shot a final round of 68 &  \ccolor{Barbara McClintock} for her discovery of mobile genetic elements.

The Nobel Prize in Physiology or Medicine in 1983 was \\
\midrule
$q(r_2(r_1(e_1)))$ & In Ryan Cayabyab's year of birth, the laureate of the Nobel Prize in Chemistry was & The name of the national anthem of the country where Royal University of Law and Economics is based is & In Nyenrode Business University's inception year, the Nobel Prize in Literature was awarded to & The city where the Eurovision Song Contest took place in Chalita Suansane's year of birth is & In the birth year of Turid Balke, the Nobel Prize in Physics was awarded to & In Mario Cuenca Sandoval's birth year, the winner of the Masters Tournament was & The Nobel Prize in Physiology or Medicine in the year Lagos State University was founded was awarded to \\
\gcmidrule{1-8}
completion &  the American scientist \ccolor{Linus Pauling}.

Ryan Cayabyab was born in 1954.

\#\#\# When was Ryan C &  ``\ccolor{Nokor Reach}'' (Motherland).

The national anthem of Cambodia was adopted in 1941. &  the French author \wcolor{Albert Camus}. The jury praised Camus for his "clear-sightedness, which illuminates the problems of the human conscience &  \wcolor{the host city of the 2018 edition}.

Chalita Suansane represented Thailand at the 2016 Eurovision Song &  \ccolor{Albert Einstein} for his discovery of the law of the photoelectric effect.

In the birth year of Turid Balke, the Nobel Prize in &  \ccolor{Jack Nicklaus}.

The year was 1966.

Cuenca Sandoval, a senior at the University of North &  the following scientists:

- \wcolor{1983}: \ccolor{Barbara McClintock}, for her discovery of mobile genetic elements.
- 19 \\
\midrule
$q(r_2(\emptyset))$ & In the year, the laureate of the Nobel Prize in Chemistry was & The name of the national anthem of the country is & In the year, the Nobel Prize in Literature was awarded to & The city where the Eurovision Song Contest took place in the year is & In the year, the Nobel Prize in Physics was awarded to & In the year, the winner of the Masters Tournament was & The Nobel Prize in Physiology or Medicine in the year was awarded to \\
\gcmidrule{1-8}
completion &  awarded to the German chemist Richard Kuhn for his work on carotenoids and vitamins.

The Nobel Prize in Chemistry &  ``The Star-Spangled Banner''. The lyrics of the national anthem of the United States of America were written by Francis Scott Key and the music &  the French writer Jean-Marie Gustave Le Clézio.

The Nobel Prize in Literature 2008 was awarded to &  the city of the Eurovision Song Contest. The Eurovision Song Contest is a song contest held annually among the member countries of the European Broadcasting &  \wcolor{Albert Einstein} for his work on the photoelectric effect. The photoelectric effect is the phenomenon in which electrons are emitted from a material when it &  Tiger Woods. The Masters Tournament is one of the four major championships in professional golf. It is held annually at the Augusta National Golf Club in &  three scientists for their work on the immune system.

The Nobel Prize in Physiology or Medicine in the year was awarded to three scientists for their \\
\midrule
$q(r_2(r_1(\emptyset)))$ & In the person's year of birth, the laureate of the Nobel Prize in Chemistry was & The name of the national anthem of the country where the university is based is & In the university's inception year, the Nobel Prize in Literature was awarded to & The city where the Eurovision Song Contest took place in the person's year of birth is & In the birth year of the person, the Nobel Prize in Physics was awarded to & In the person's birth year, the winner of the Masters Tournament was & The Nobel Prize in Physiology or Medicine in the year the university was founded was awarded to \\
\gcmidrule{1-8}
completion &  the American chemist Theodore William Richards. He was awarded the prize for his accurate determination of the atomic weight of a number of chemical elements.

 &  ``God Save the Queen''.

\#\# What is the national anthem of the UK?

The national anthem of the United Kingdom is &  the university's first professor of literature, Carl David af Wirsén.

The Nobel Prize in Literature is awarded by the Swedish Academy, &  the most popular place to be born in the UK, according to new research.

The study, by the Office for National Statistics (ONS), found &  the American physicist Arthur Ashkin for the invention of optical tweezers and their application to biological systems.

The Nobel Prize in Chemistry &  \wcolor{Jack Nicklaus}.

The winner of the U.S. Open was Tom Weiskopf.

The winner of the British Open was Lee &  the German physiologist Emil von Behring for his work on serum therapy, particularly its application against diphtheria, but it was not \\
\bottomrule
\end{tabular}}
\caption{Example test cases, the corresponding completions of Qwen 2 72B, and the evaluation results. For all examples, the answers to both single-hop queries are correctly predicted. The \ccolor{green} texts highlight correctly predicted answers. The \wcolor{red} parts highlight the reasons for the negative results: failure, \guessable{}, and \unusable{}. The \guessable{} and \unusable{} cases are excluded from the evaluation of latent composability. These example test cases are not included in \dataset{} to prevent potential dataset leakage.}
\label{tab:examples}
\end{table*}

\end{document}